\UseRawInputEncoding
\documentclass[preprint,12pt]{elsarticle}

\usepackage{graphicx}
\usepackage{amssymb} 
\usepackage{epstopdf}
\usepackage{listings}
\usepackage{float}
\usepackage{lineno}
\usepackage{etex}
\usepackage{graphicx,color,colordvi}%subfigure
\usepackage{latexsym,amsbsy,epsfig,fancybox}
\usepackage{amstext}
\usepackage{subfigure}
\usepackage{amsfonts,textgreek}
\usepackage{mathrsfs}
\usepackage{mathtools}
\usepackage{stmaryrd,dsfont}
\usepackage{bbm}
\usepackage{url}
\usepackage{wrapfig,bm}
\usepackage{tikz}
\usepackage{pstricks}
\usepackage{caption}
\usepackage{psfrag}
\usepackage{upgreek}
\usepackage{isomath}
\usepackage{latexsym}
\usepackage{array}
\usepackage{multirow}
\usepackage{booktabs}
\usepackage{verbatim}
\usepackage{natbib}
\biboptions{sort&compress}
\usepackage{color}
\usepackage{colortbl}
\usepackage[colorlinks=true,citecolor=blue]{hyperref}
\usepackage{overpic}
\usepackage{enumerate}
\usepackage{amsmath,amssymb,xspace}
\usetikzlibrary{plotmarks}
\usetikzlibrary{fit,positioning,arrows}
\usetikzlibrary{shapes.geometric, plotmarks}
\usepackage{appendix}

\usepackage[margin=3cm]{geometry}
\usepackage{soul}

\newcommand{\ud}{{\rm d}}
\newcommand{\ui}{{\rm i}}

\tikzstyle{nicebox}=[draw=black!100, fill=white!10, rectangle, inner sep=4pt, inner ysep=16pt]
\tikzstyle{niceboxtitle}=[draw=black!100, fill=white, text=black, rectangle]

\usetikzlibrary{arrows,decorations.markings}

\journal{Computer Physics Communications}
\begin{document}
\begin{frontmatter}
\title{Evaluating the Transferability of Machine-Learned Force Fields for Material Property Modeling}
\author[1]{Shaswat Mohanty\fnref{label1}} 
\author[2]{SangHyuk Yoo\fnref{label1}} 
\author[2]{Keonwook Kang} 
\author[1]{Wei Cai\corref{cor1}}
\ead{caiwei@stanford.edu}
\cortext[cor1]{Corresponding author}
	
\address[1]{Department of Mechanical Engineering, Stanford University, CA 94305-4040, USA}
\address[2]{School of Mechanical Engineering, Yonsei University, Seoul 03722, Republic of Korea}
\fntext[label1]{Equal contribution}

\begin{abstract}
%% Text of abstract
Machine-learned force fields have generated significant interest in recent years as a tool for molecular dynamics (MD) simulations, with the aim of developing accurate and efficient models that can replace classical interatomic potentials.
However, before these models can be confidently applied to materials simulations, they must be thoroughly tested and validated.
The existing tests on the radial distribution function and mean-squared displacements are insufficient in assessing the transferability of these models.
Here we present a more comprehensive set of benchmarking tests for evaluating the transferability of machine-learned force fields.
We use a graph neural network (GNN)-based force field coupled with the OpenMM package to carry out MD simulations for Argon as a test case. Our tests include computational X-ray photon correlation spectroscopy (XPCS) signals, which capture the density fluctuation at various length scales in the liquid phase, as well as phonon density-of-state in the solid phase and the liquid-solid phase transition behavior.
Our results show that the model can accurately capture the behavior of the solid phase only when the configurations from the solid phase are included in the training dataset.  This underscores the importance of appropriately selecting the training data set when developing machine-learned force fields.
The tests presented in this work provide a necessary foundation for the development and application of machine-learned force fields for materials simulations.
\end{abstract}


\end{frontmatter}
\section*{Highlights}
\begin{itemize}
    \item Development of a suite of benchmarking tests, including X-ray photon correlation spectroscopy for machine-learned force fields (code available).
    \item Phonon density of states and melting point calculations are necessary tests for the solid phase.
    \item Training data should include both solid and liquid configurations to accurately model material behavior.
\end{itemize}
\section{Introduction} 
\label{sec:Intro}

Molecular dynamics (MD) is a powerful tool for studying the equilibrium and transport properties of many-body systems with applications in diverse fields such as materials science~\cite{agrawal2019deep,choudhary2022recent,huang2020practicing}, polymer chemistry~\cite{ethier2021deep,chandrasekaran2020deep,shayeganfar2021deep}, biochemistry~\cite{sino2021review} and medical science~\cite{bhatt2021state,suzuki2017overview,shen2017deep}.
In MD simulations, the motion of the atoms is driven by interatomic forces, which are evaluated using interatomic potentials.
However, there are limitations to traditional molecular dynamics simulations, including a limited length scale, a limited time scale, and insufficient force field accuracy.
In recent years, machine learning techniques have been used to address these limitations, with the goal of accelerating molecular dynamics simulations and improving force field accuracy. 
Machine learning has been used to predict atomic trajectories~\cite{wang2018deepmd,zheng2021learning} with the ultimate aim of bridging the timescale gap between MD simulations and experiments~\cite{do2022glow,perez2009accelerated}. 
However, these techniques often require retraining the 
model for different system sizes, making it challenging to apply to large systems, which are often necessary for property predictions.
As a result, there is significant interest in developing scalable machine-learning methods for MD simulations.

Graph-based deep learning techniques have proven effective in developing accurate and scalable force fields that can be trained on data from small simulation cells and then applied to larger systems.
Graph neural networks (GNN) can accurately predict atomic forces with a computational cost that scales linearly with the number of atoms, making them well-suited for large-scale simulations.
Recently, GNN-based force fields have been used to predict MD trajectories using approaches such as 
SchNet~\cite{schutt2018schnet},
and DeepMD~\cite{wang2018deepmd,marcolongo2020simulating,balyakin2020machine}.
These force fields, when trained on high-quality \emph{ab initio} data, can perform large-scale simulations at an accuracy comparable to that of \emph{ab initio} model but only at a small fraction of the cost~\cite{smith2020ani,dajnowicz2022high,mohanty2023development}. However, before GNN force fields (or any machine-learned force fields) can be confidently deployed, their transferability to configurations beyond the training data set must be established. Currently, most existing benchmark tests on machine-learned force fields are limited to the radial distribution function and self-diffusivity in the liquid phase. As we show below, these tests are insufficient for establishing their reliability for materials modeling.

To this end, we present a series of benchmarking tests, comparing the predictions of GNN-based force fields~\cite{li2022graph} to those of the original model that produces the training data.
As an example, the original model is the classical Lennard-Jones interatomic potential for Argon.
The purpose of choosing such a simple model as a benchmark in this work is to provide a baseline to evaluate the transferability of machine-learned force fields without incurring significant computational costs.
Using such a simple model, a large amount of data can be easily generated for training and comparison purposes.  The findings from this study pave the road for future tests for machine-learned force fields fitted to \emph{ab initio} data which are much more expensive to generate. 
In addition to the radial distribution function and self-diffusivity, we 
compare the computational X-ray photon correlation spectroscopy (XPCS) signals~\cite{Mohanty2022}, which capture density fluctuations at different length scales in the liquid phase.
We further show that a model trained exclusively on liquid configurations 
fails to accurately capture the vibrational frequency distribution in the solid phase. 
This deficiency is fixed only after the model is trained on configurations sampled from both the liquid and solid phases.
These findings raise a concern about the transferability of the GNN-based force field and underscore the importance of ensuring that the training data adequately cover the areas of interest in the configurations space.

The paper is structured as follows. In Section~\ref{sec:Model_form} we discuss the preparation of training data, the GNN model details, and the hyperparameters used for training the model. Additionally, we present the metrics that will be used to analyze the performance of the GNN-MD simulations against the MD simulations using the original model. In Section~\ref{sec:Numerical}, we present the results of these benchmark tests.
We 
summarize our findings in Section~\ref{sec:conclusion}.

\section{Model details}
\label{sec:Model_form}
GNN-based models provide a linearly scalable force field that has the potential to be used in \emph{ab initio} molecular dynamics (AIMD) simulations. However, verifying a model against direct \emph{ab initio} simulations are computationally expensive, especially for larger systems. To address this issue, we will focus on conducting benchmarking tests on a GNN force field that is trained on forces generated using the classical Lennard-Jones interatomic potential.
This will allow us to analyze the model's performance without the added computational cost of using \emph{ab initio} data.
We assume that the transferability of the GNN model can be assessed when trained on forces from either classical interatomic potentials or \emph{ab initio} data, with comparable results.
\subsection{Preparation of reference data} \label{sec:subsec:prep_ref}
We prepared a 256-atom configuration of Argon (Ar), described by the Lennard-Jones (LJ) interatomic potential used in our earlier work~\cite{Mohanty2022}, by using \texttt{LAMMPS}~\cite{LAMMPS}. 
\begin{align}
    V_{\rm LJ}(r) &=4 \epsilon\left[ \left(\frac{\sigma}{r}\right)^{12}-\left( \frac{\sigma}{r}\right)^{6}\right],
\end{align}
where $\epsilon=0.0103$ eV, $\sigma=3.40$ \AA. The density of the system used is $\rho$ = 0.858 amu/\AA$^{3}$ (0.844 in LJ unit system). The cut-off radius of the neighbor list is 8.5 \AA \, (2.5 in LJ unit system).
Periodic Boundary Conditions (PBC) were applied in all three directions. The initial positions of atoms were generated according to a perfect face-centered cubic (FCC) lattice and the velocities were assigned random values corresponding to a temperature of 10~K. To reach the state of thermal equilibrium, the Nosé-Hoover NPT ensemble and the NVT ensemble were applied sequentially. The equilibration simulation was performed for about 2.156~ns, with a timestep of 10.78~fs. Simulations under both ensembles used a thermostat with a collision frequency of 0.02 /ps, a chain length of 5, and 5 MTK~\cite{martyna1999reciprocal} loops were used. For pressure control under the NPT ensemble, a barostat with a collision frequency of 0.2 /ps was used. After implementing this protocol of \emph{initial equilibration}, the thermostat temperature (NVT simulation) was increased with a linear ramp from 10~K to 105~K over the 2.156~ns trajectory, during which atomic configurations were saved  every 100 timesteps. This same process is repeated ten times with different initialization of velocities to generate a total of 20000 configurations.
Additionally, the configuration at the end of the \emph{initial equilibration} protocol is used as the initial state to run the analysis trajectory by using \texttt{LAMMPS} for the MD results and the \texttt{Atomic Simulation Environment}  \cite{ASE} and \texttt{OpenMM} \cite{OpenMM} script for the GNN-MD simulation, respectively.
\subsection{GNN model}
Our benchmarking tests are carried out on the GNN-based force field following the GNN structure and training procedure of \citet{li2022graph}. For completeness, we present a brief description of the GNN model in this section. For the GNN-based force predictor, the atomic configurations are first converted into graphs that encode the local atomic environment. Here, each atom acts as a node, and the vectors towards neighboring atoms within a cut-off radius act as edges in the graph. 
The node information includes the atomic position and force vector. 
The edge information includes the distance between neighboring atoms and the unit vector toward the neighbors. A brief description of each step in the GNN force field prediction is enlisted below. A more detailed description of the model can be found in \cite{li2022graph}.
\begin{itemize}
    \item \emph{Encoding layer} \\
    The feature $v_{i}^{(l)}$ is an array of size $h$ and represents the encoded node feature at the $l^{\rm th}$ message-passing layer. The reference node encoding, $v_{i}^{(0)}$, is created by passing the atomic species information $s_{i}$ through an encoding multi-layer perceptron (MLP), $e^N$, such that $v_{i}^{(0)}=e^N(s_{i})$. 
    The feature $e_{ij}^{(l)}$ is also an array of size $h$ represents the encoded edge feature at the $l^{\rm th}$ message-passing layer. The reference edge feature is also created through an MLP, $e^E$, where $e_{ij}^{(l)}=e^E(\mathbf{q}_{ij},d_{ij})$. Here, for a given pair of atoms[$i$,$j$], the unit vector and the distance between them are denoted by $\mathbf{q}_{ij}$ and $d_{ij}$, respectively.
    The superscript $l$ is the index of the message-passing layer, 
    %. 
    which varies from 0 to $n$ and denotes how many layers the features have passed through. 
    At the encoding step, $d_{ij}$ is represented by the radial basis functions, following the implementation of Schnet~\cite{schutt2018schnet}. 
    \item \emph{Message passing layer} \\
    The encoded features, $v_{i}^{(l)}$ and $e_{ij}^{(l)}$, are passed through the message passing layer $m_{j\rightarrow{i}}^{(l)}$. The equation of the message-passing layer is as follows,
    \begin{equation}
        m_{j\rightarrow{i}}^{(l)} = \Phi^{(l)}\left(v_{j}^{(l-1)} + e_{ij}^{(l-1)} + v_{i}^{(l-1)}\right)\odot v_{j}^{(l-1)},\qquad \forall j\in \mathcal{N}(i),
    \end{equation}
    where $\Phi^{(l)}$ denotes a MLP with the activation function being the Gaussian Error Linear Units (GELU) function \cite{gelu}. As an input of $\Phi^{(l)}$, the edge features and node features are added up as a vector sum since their dimensions are the same. $\odot$ refers to elemental multiplication, and $j$ is the index of the set of neighbors $\mathcal{N}(i)$ of the $i^{\rm th}$ node. 
    \\
    Each message from $j$-th atom aggregates into $M_{i}^{(l)}$ and the new node feature $v_{i}^{(l)}$ is computed by the following equations.
    \begin{align}
        M_{i}^{(l)} =& \sum_{\forall j\in \mathcal{N}(i)} m_{j\rightarrow{i}}^{(l)}, \\
        v_{i}^{(l)} =&\, \Theta^{(l)}\left(v_{i}^{(l-1)} + M_{i}^{(l)}\right) + v_{i}^{(l-1)}.
        \label{eq:vil_update}
    \end{align}
    Here, $\Theta^{(l)}$ is the node update MLP function at the $l^{\rm th}$ layer. 
    Eq.~(\ref{eq:vil_update}) shows that the 
    node features are updated recursively.
    However, the edge features are not updated recursively~\cite{li2022graph}.
    Instead, the edge features at every level are computed directly from the reference edge encoding through an MLP function, $A^{(l)}$, i.e.,
    $e_{ij}^{(l)}=A^{(l)}(e_{ij}^{(0)})$.
    \item \emph{Decoding layer} \\
    At the last step of the neural network, the updated node feature is decoded into the interatomic forces $\bm{f}_{i}$ by executing a forward pass through the decoding MLP.
\end{itemize}
Although we present our benchmarking tests on a model trained on pair potential, the GNN model is not constrained to a force field derived from a pair potential form. The GNN architecture used here is also capable of learning a force field derived from \emph{ab initio} data~\cite{li2022graph}.
\subsection{Details of the training}
We set the number of encoding features and latent features as $h = 128$ and used $n = 4$ message-passing layers in the GNN. The cut-off radius for converting atomic configuration to graph is 7.5 \AA, which is 1 \AA\  smaller than that in the LJ potential. The effects of the cutoff radius will be studied in the future.
The loss function for the training contains the L1 distance between the reference force $\bm{f}_{i}$ and the predicted force $\hat{\bm{f}_{i}}$, as well as a penalty function, as follows.
\begin{equation}
    \mathcal{L}=\frac{1}{N}\sum_{i=1}^{N}\left\lVert\bm{f}_{i}-\hat{\bm{f}_{i}}\right\rVert_1 + \lambda \left\lVert\frac{1}{N}\sum_{i=1}^{N}\hat{\bm{f}_{i}}\right\rVert_1,    
\end{equation}
where $N$ is the number of atoms in the reference data. The force on the center of mass from an interatomic potential is expected to be zero. However, the GNN model does not compute the forces from the derivative of a potential function, we are not guaranteed a zero force on the center of mass. We minimize the magnitude of the total force on the entire system to prevent divergent dynamics of the center of mass by adding a penalty function to the loss function. The regularization parameter, $\lambda$, is set to 0.01 during the training.
As described earlier, we used 20,000 atomic configurations as our training dataset. The dataset is shuffled and split into the train and test set with a 90:10 ratio. During training, we carry out standard normalization of the atomic forces such that the mean atomic force is zero and the variance is unity.
We used \texttt{PyTorch 1.11.0}, \texttt{PyTorch-lightning 1.6.3}, \texttt{DGL 0.8.1} and \texttt{Scikit-learn 0.24.2} packages to train and test the model. We use the Adam optimizer with an exponential learning rate scheme (from $3\times10^{-4}$ to $1\times10^{-7}$) during the training of the force field. The total number of epochs is set to 30 and we choose the trained model parameter (weights and biases) at the last epoch as the force calculator for our GNN-MD simulation.
\subsection{Metrics for model performance}
Here we list the static and dynamic metrics that will be used to compare the performance of the GNN-MD simulations against the MD simulations.  To examine the structure of the liquid phase we will study the pair distribution function.
To examine the dynamics of the liquid phase, we evaluate the self-diffusivity and carry out the computational XPCS analysis on the MD trajectory. 
The relations to compute the properties such as the radial distribution function, the structure factor, and the computational XPCS signal, are described only briefly in \ref{subsec:RDF} and \ref{subsec:XPCS} (see \cite{Mohanty2022} for more details), since the focus of this paper is to establish the accuracy of the GNN-MD in capturing the dynamics and structure of the simulation.
To evaluate the model performance in the solid phase, 
%we perform melting simulations. In addition
we compute the phonon density of states (PDOS), and the melting point by using the interface method. Further details on the key equations we solve to obtain the numerical results are given in \ref{app:model}.

\section{Numerical Results} 
\label{sec:Numerical}
The MD and GNN-MD simulations discussed in this section are carried out on a 256-atom and a 4000-atom system to test the scalability and transferability of the GNN force field which was trained on 256-atom configurations. The training samples for the 256 and 4000 atoms system are obtained from MD simulations using \texttt{LAMMPS}~\cite{LAMMPS}. 
While in the previous work~\cite{li2022graph} only liquid configurations are included in the training data, here we train two models:
Model A is trained only using liquid configurations and Model B is trained using both liquid and solid configurations (see Section~\ref{sec:subsec:prep_ref} for the procedure of preparing these configurations).
The results presented in Sections~\ref{sec:subsec:acc_force},~\ref{sec:subsec:str_liq}, and \ref{sec:subsec:dyn_liq}, corresponds to Model B. However, Model A gives similar results for the tests in these sections; hence they are omitted here for brevity. 
On the other hand, Model A and Model B give different results for some of the tests involving the solid phase, and both will be presented in Section~\ref{sec:subsec:sol_phase}.

The initial configuration for the testing MD and GNN-MD simulations are prepared using the \emph{initial relaxation} protocol.
 The final trajectory for the analysis is then run using the same NVT ensemble as in the final relaxation step. However, the total simulation time extends to 539 ps with a timestep of 10.78 fs. The simulation frames are stored at 107.8 fs intervals for our analysis. In comparison to MD simulations using the LJ potential, the GNN-MD is $10^1$ times slower, because the atomic forces from the LJ potential can be computed very quickly. We reiterate that the focus of this work is to use the LJ potential as an example to test the transferability of the GNN force field in capturing the static and dynamic properties of the material system that it is trained on.
\subsection{Accuracy of force prediction} \label{sec:subsec:acc_force}
\begin{figure}[H]
    \centering
    \subfigure[]{\includegraphics[width=0.31\textwidth]{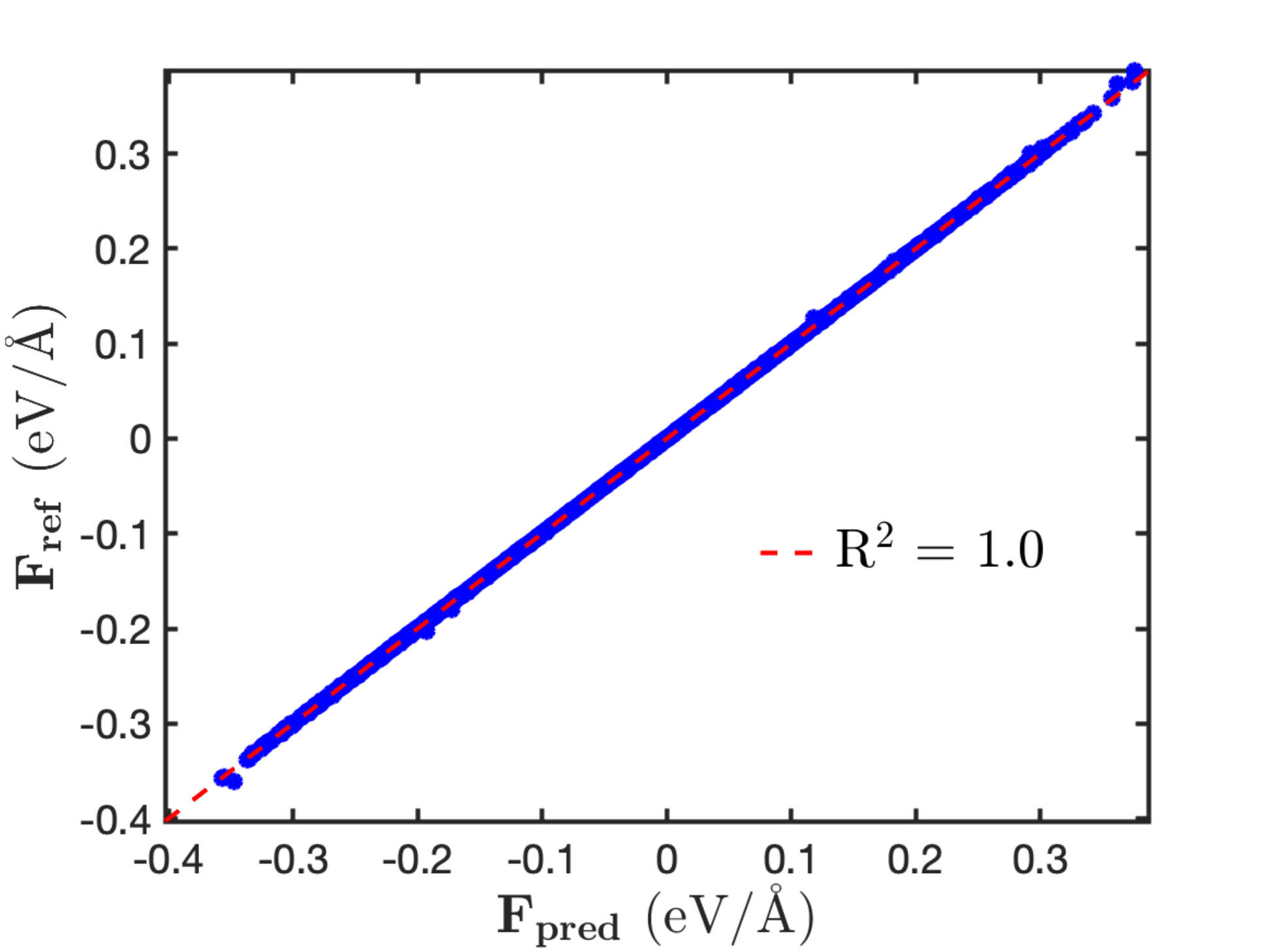}
    \label{fig:pred_F}}
    \subfigure[]{\includegraphics[width=0.31\textwidth]{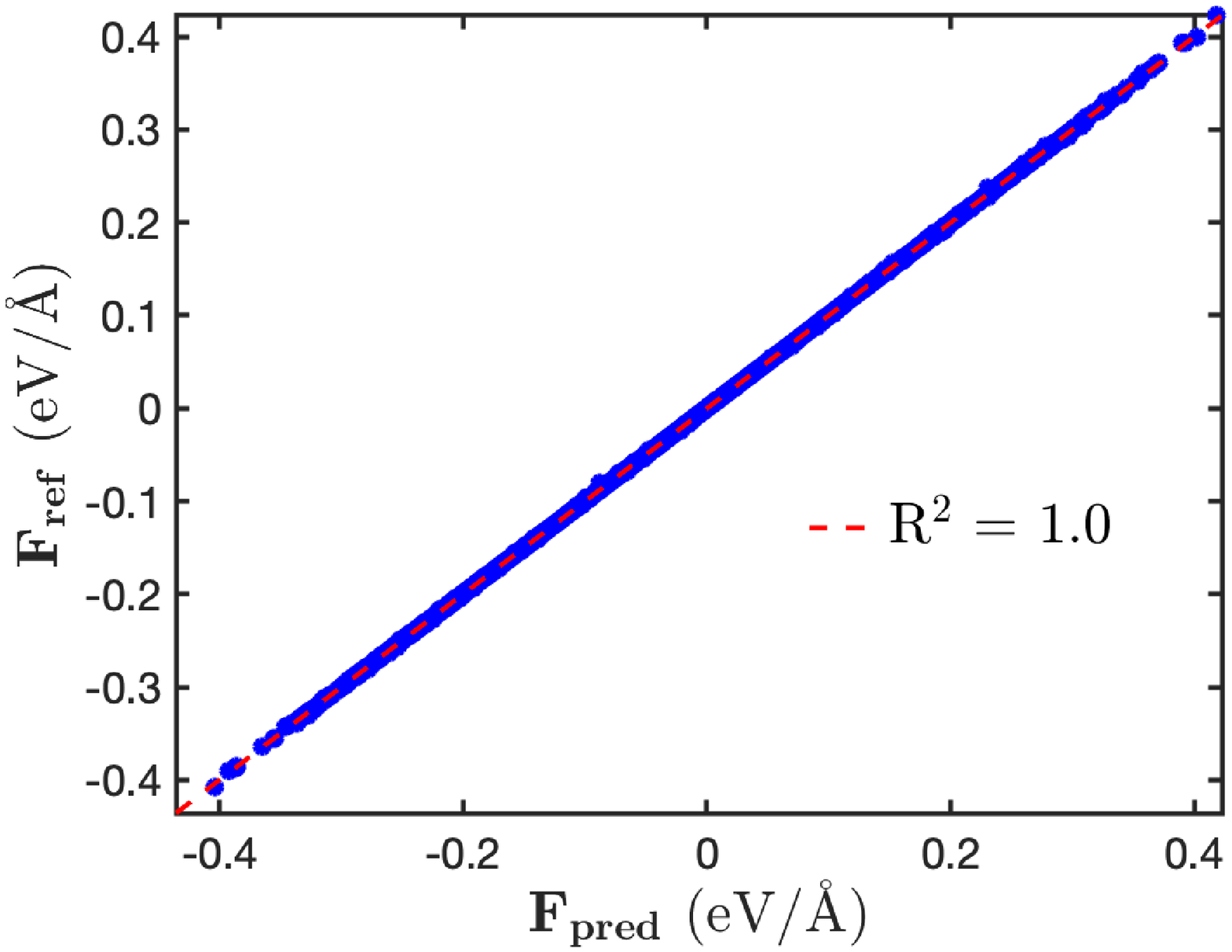}
    \label{fig:pred_F_4000}}
    \subfigure[]{\includegraphics[width=0.31\textwidth]{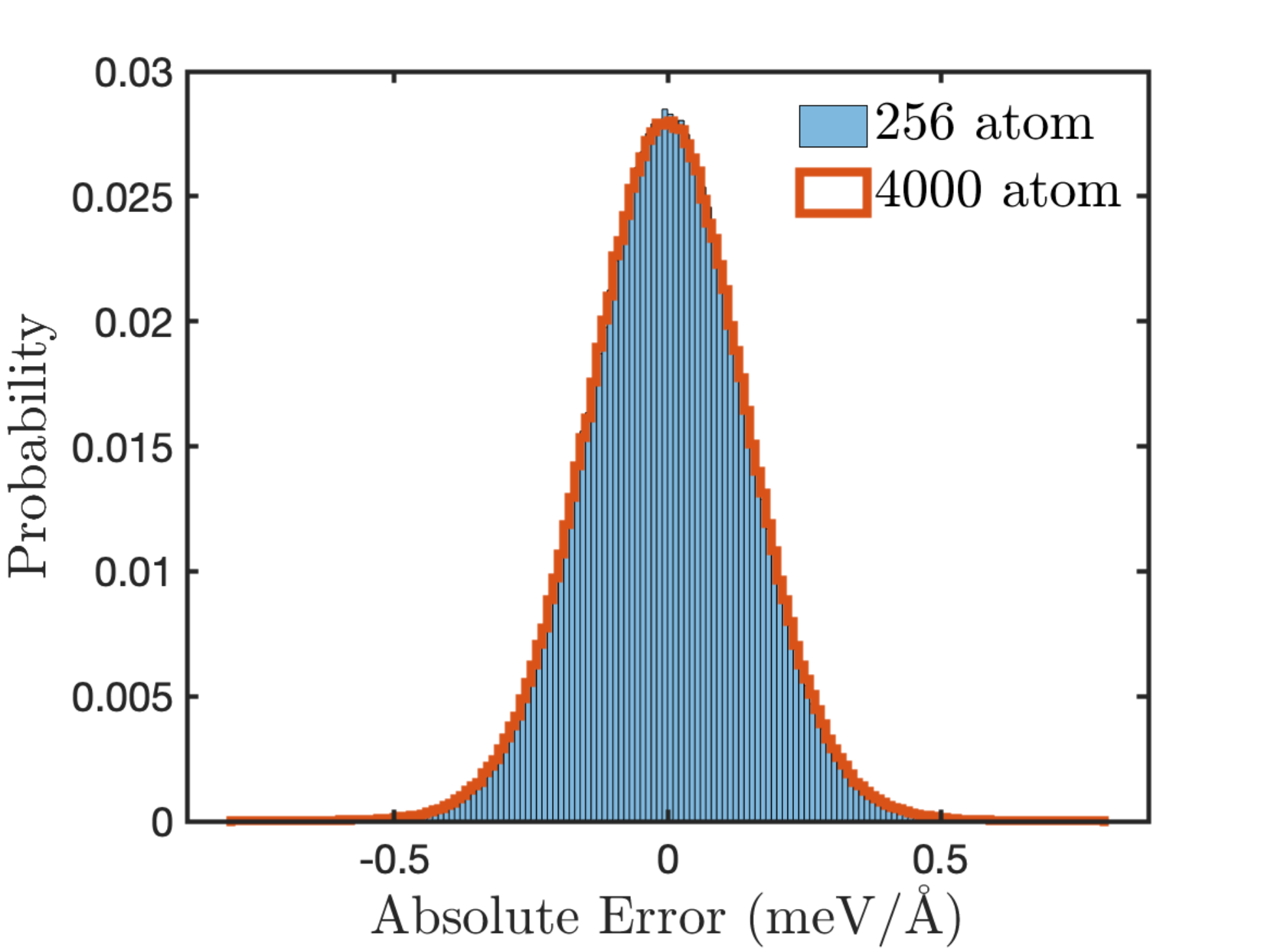}
    \label{fig:MAE}}
    \caption{Parity plot for the predicted forces and the actual forces for the (a) 256 atom and (b) 4000 atom system. (c) Distribution of the absolute error of the predicted forces for the 256 atom and 4000 atom system over the configurations of analysis.  \label{fig:model_perf}}
\end{figure}

We test the atomic forces predicted atomic forces against the calculated atomic forces from the Lennard-Jones potential for 1,000 sampled configurations of our 256-atom simulation. These configurations are different from the ones used for the training of the GNN force field. We see that the predicted forces are well correlated with the actual forces across all atoms and all configurations, as shown in Fig.~\ref{fig:pred_F} (parity plot). The coefficient of determination, R$^2$, is almost $1.0$ ($1 - $R$^2 \approx 10^{-4}$) which shows how well correlated the predicted force is to the actual forces computed from LAMMPS by using the Lennard-Jones force field. The generalizability of the force field to a 4000-atom system is captured by the parity plot shown in Fig.~\ref{fig:pred_F_4000}, where the quality of the force prediction remains qualitatively the same.
Figure~\ref{fig:MAE} shows the distribution of the absolute error of the predicted forces,
for the 256-atom (distribution of atomic force components over 1,000 configurations) and the 4000-atom (distribution of atomic force components over 100 configurations) system. 
The standard deviation of the force error is 0.15 meV/\AA.
We see that the absolute error distribution is the same irrespective of the system size, which indicates the error of the trained model and is independent of the size of the system. This establishes the quality of the trained model consistent with previous report~\cite{li2022graph}.

\subsection{Structure of liquid phase} \label{sec:subsec:str_liq}
\begin{figure}[H]
    \centering
    \subfigure[]{\includegraphics[width=0.48\textwidth]{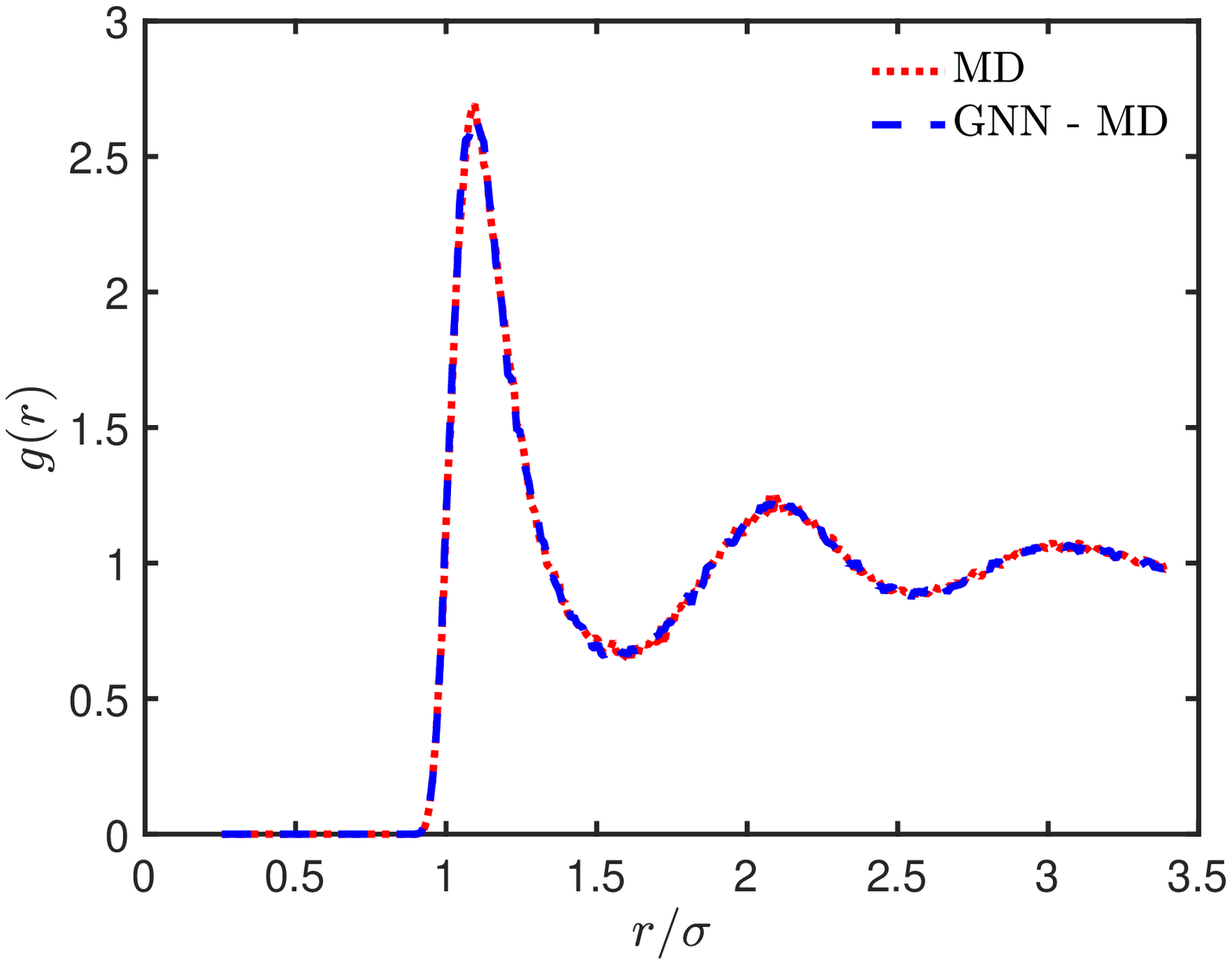}
    \label{fig:gr_256}}
    \subfigure[]{\includegraphics[width=0.48\textwidth]{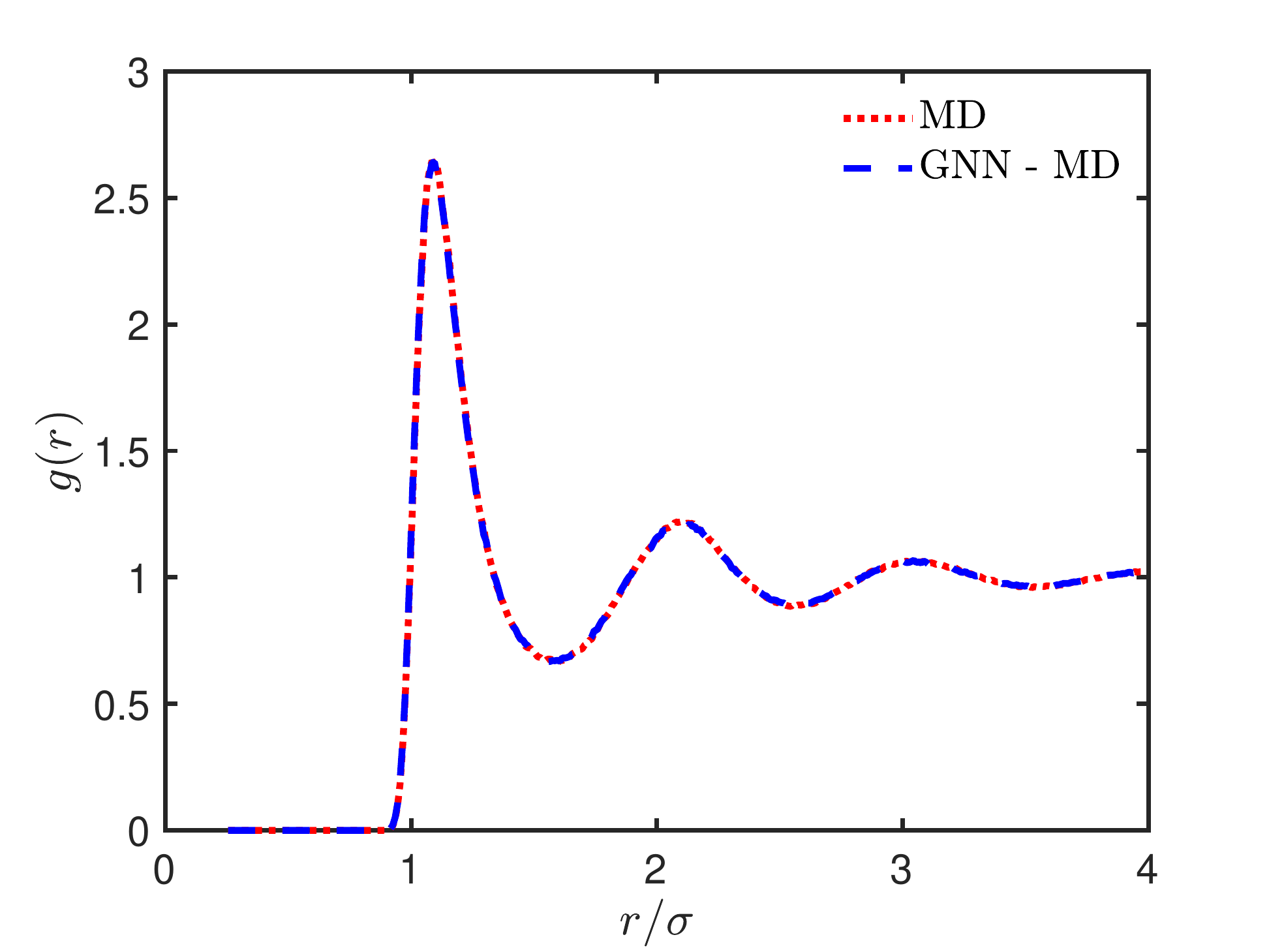}
    \label{fig:gr_4000}}
    \caption{The $g(r)$ over 100 configurations for the (a) 256 atoms and (b) 4000 atoms MD and GNN-MD simulation.  \label{fig:gr_all}}
\end{figure}

Here we examine the orientation-averaged radial distribution function, $g(r)$, 
as defined in \ref{subsec:RDF}. 
We average the $g(r)$ 
over 100 configurations that are sampled every 5.34 ps from our simulation. Figure~\ref{fig:gr_256} shows that the $g(r)$ obtained from the MD and GNN-MD simulation are in close agreement for the (small) system it is trained on, confirming the previous observation~\cite{li2022graph}. We further find that $g(r)$ between the MD and GNN-MD simulation agree well with each other even for the larger simulation cell containing 4000 atoms, as shown in Fig.~\ref{fig:gr_4000}. We also observe that the orientation-averaged structure factor, $S(q)$ (not shown here), which can be obtained from the Fourier transform of $g(r)$, agrees between MD and GNN-MD simulations, for both the 256-atom and 4000-atom systems.
\subsection{Dynamics of liquid phase} \label{sec:subsec:dyn_liq}
\begin{figure}[H]
    \centering
    \subfigure[]{\includegraphics[width=0.48\textwidth]{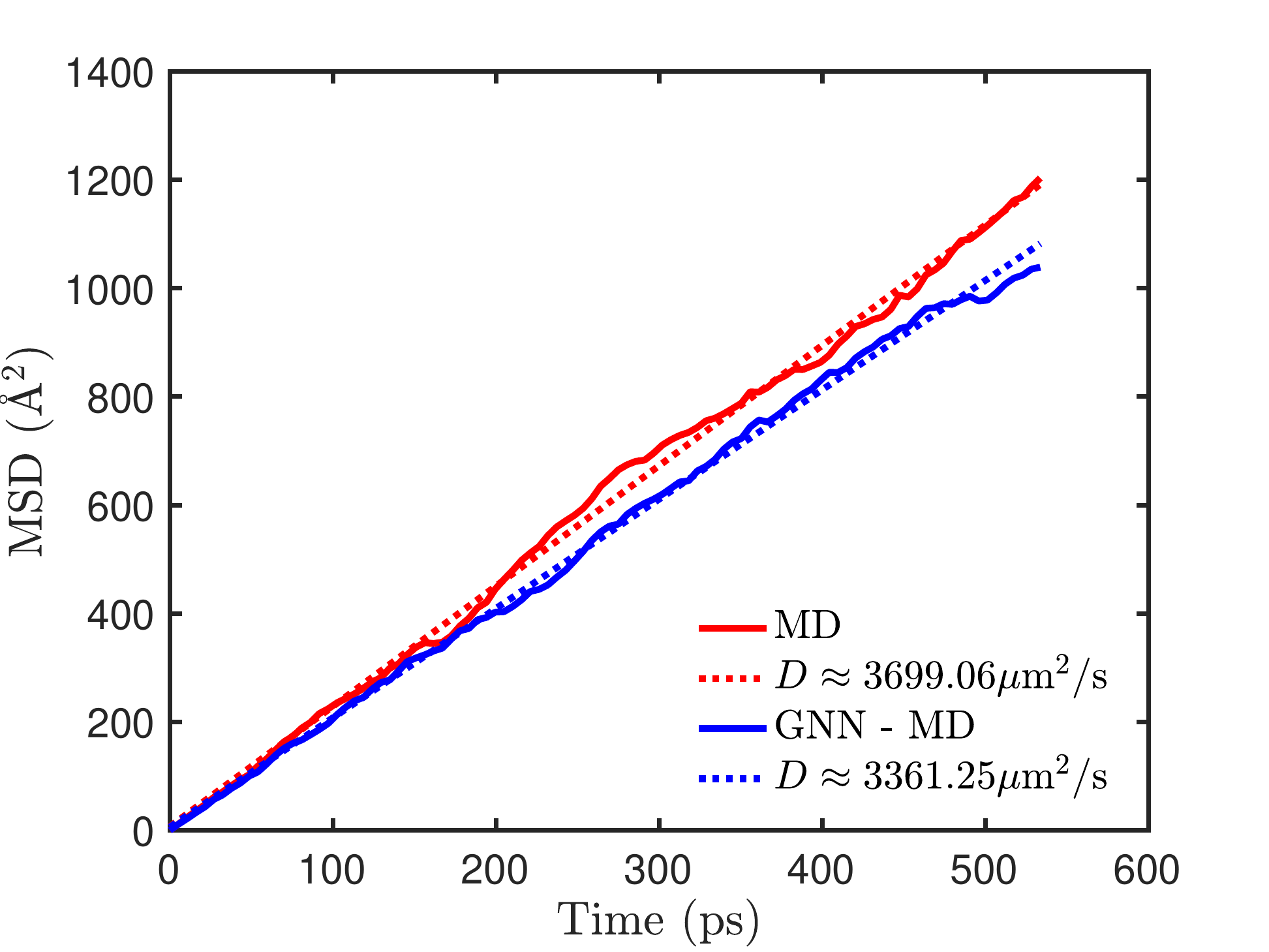}
    \label{fig:msd_256}}
    \subfigure[]{\includegraphics[width=0.48\textwidth]{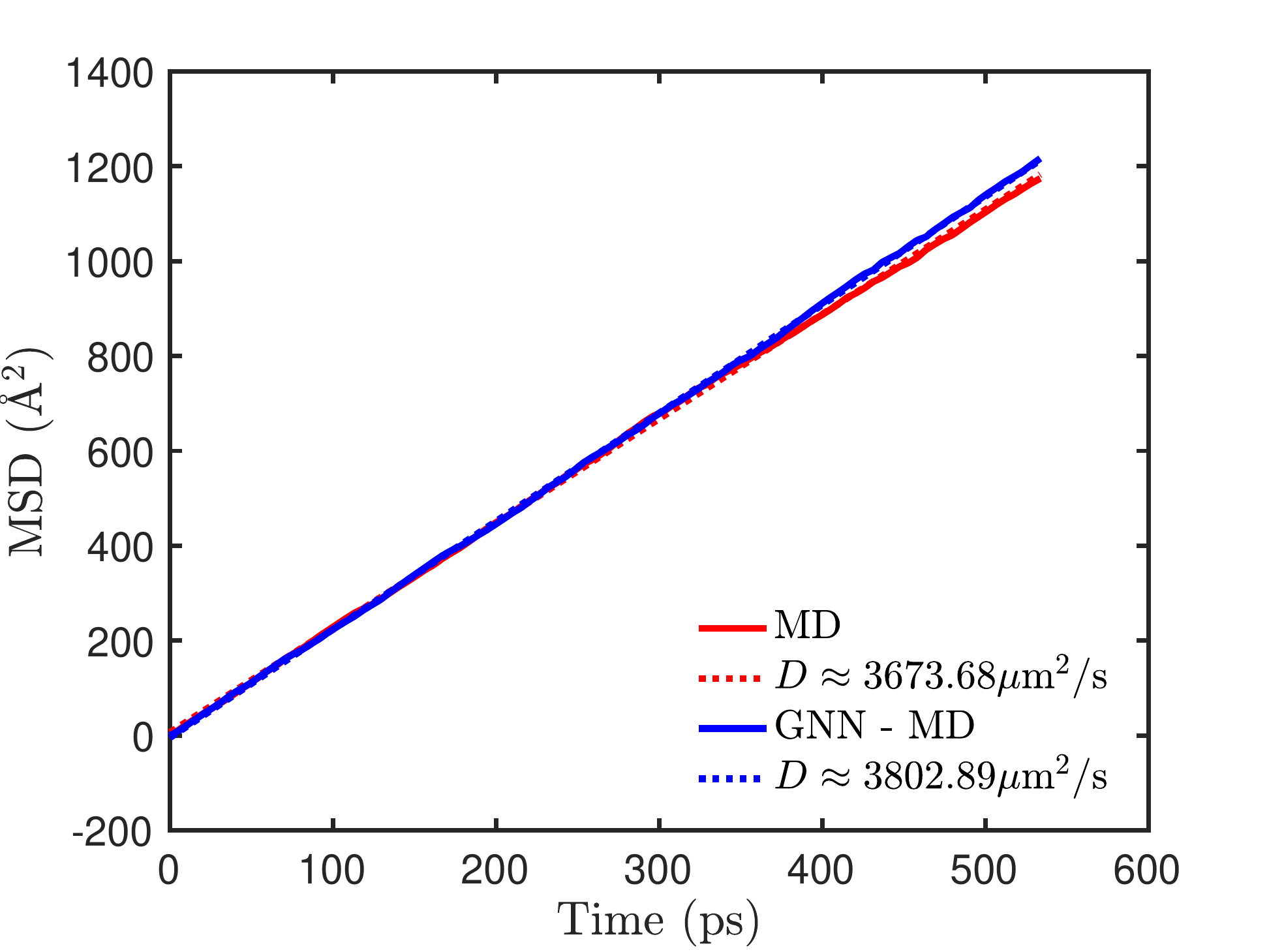}
    \label{fig:msd_4000}}
    \caption{The mean-squared displacement (MSD) for the (a) 256 atoms and the  (b) 4000 atoms from the MD and GNN-MD simulation over the 539 ps trajectory at 100 K.  \label{fig:msd_all}}
\end{figure}

Given that atomic trajectories in MD simulations are chaotic, small differences in atomic forces can lead to large divergences in the atomic trajectories at a later time.
Therefore, we don't expect the atomic trajectories in MD and GNN-MD to agree well with each other.
However, for the GNN-based force field to be useful, the statistical properties of the atomic trajectories need to be consistent with those of the original MD model.
A commonly tested statistical property
is the self-diffusivity obtained from the mean-squared displacement (MSD) as described in \ref{app:model}. 
Figure~\ref{fig:model_perf} shows that the MSD predicted by the GNN model is consistent with that from the original MD model. 
For the 256-atom system, the MSD from GNN-MD agrees very well with the MD model for the first $150$~ps, after which the difference grows somewhat larger, as shown in Fig.~\ref{fig:msd_256}.
For the 4000-atom system, the MSD from GNN-MD agrees very well with the MD model for the first $300$~ps, after which the difference also grows somewhat larger, as shown in Fig.~\ref{fig:msd_4000}.

\begin{figure}[H]
    \centering
    \subfigure[]{\includegraphics[width=0.48\textwidth]{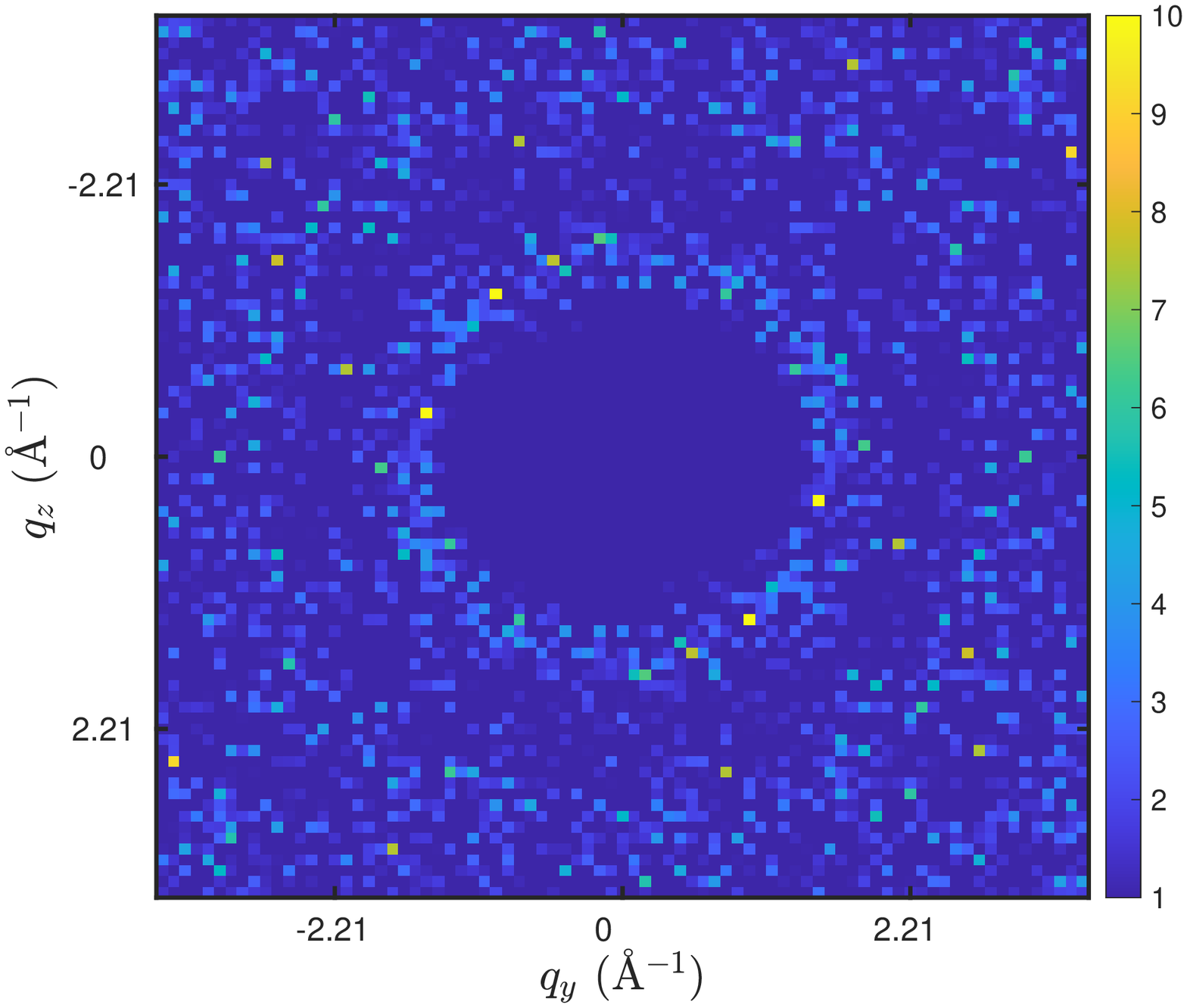}
    \label{fig:slice_MD}}
    \subfigure[]{\includegraphics[width=0.48\textwidth]{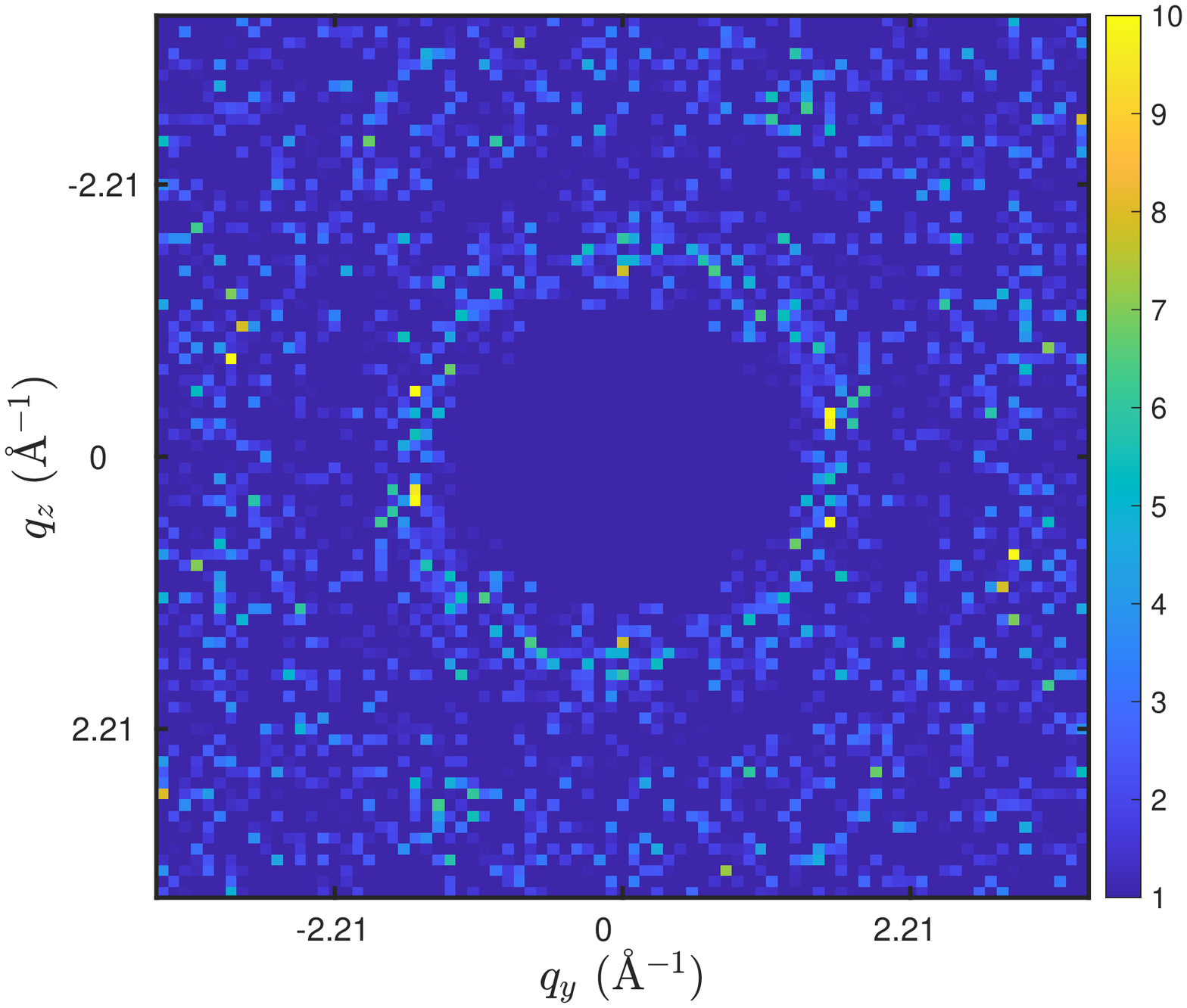}
    \label{fig:slice_GNN}}
     \caption{The speckle patterns of scattered intensity $I(\bm{q})$ on a spherical slice ($81 \times 81$ pixels) in the $\bm{q}$-space for (a) MD and (b) GNN-MD simulation.  \label{fig:slices}}
\end{figure}

\begin{figure}[H]
    \centering
    \subfigure[]{\includegraphics[width=0.48\textwidth]{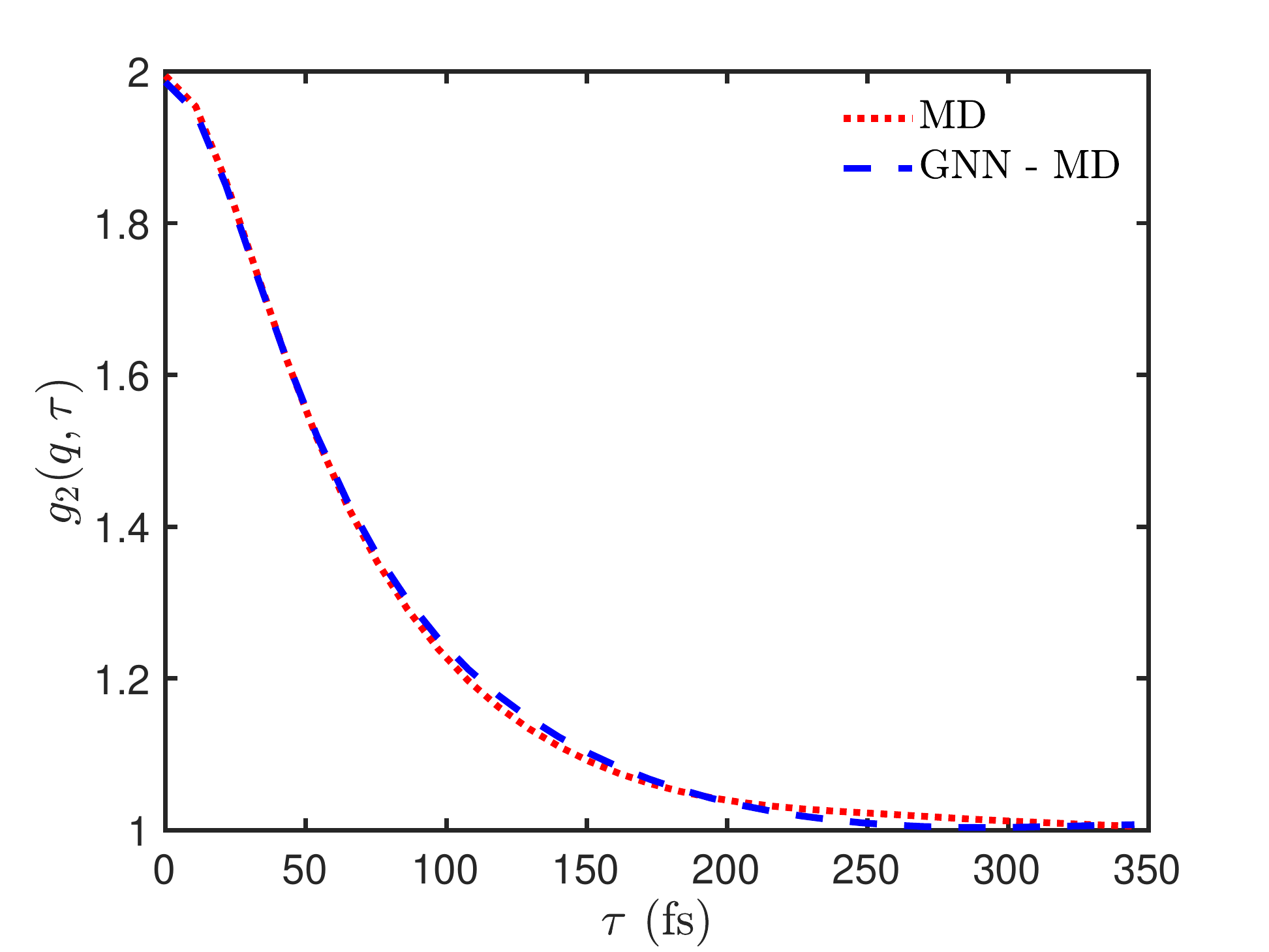}
    \label{fig:g2_256}}
    \subfigure[]{\includegraphics[width=0.48\textwidth]{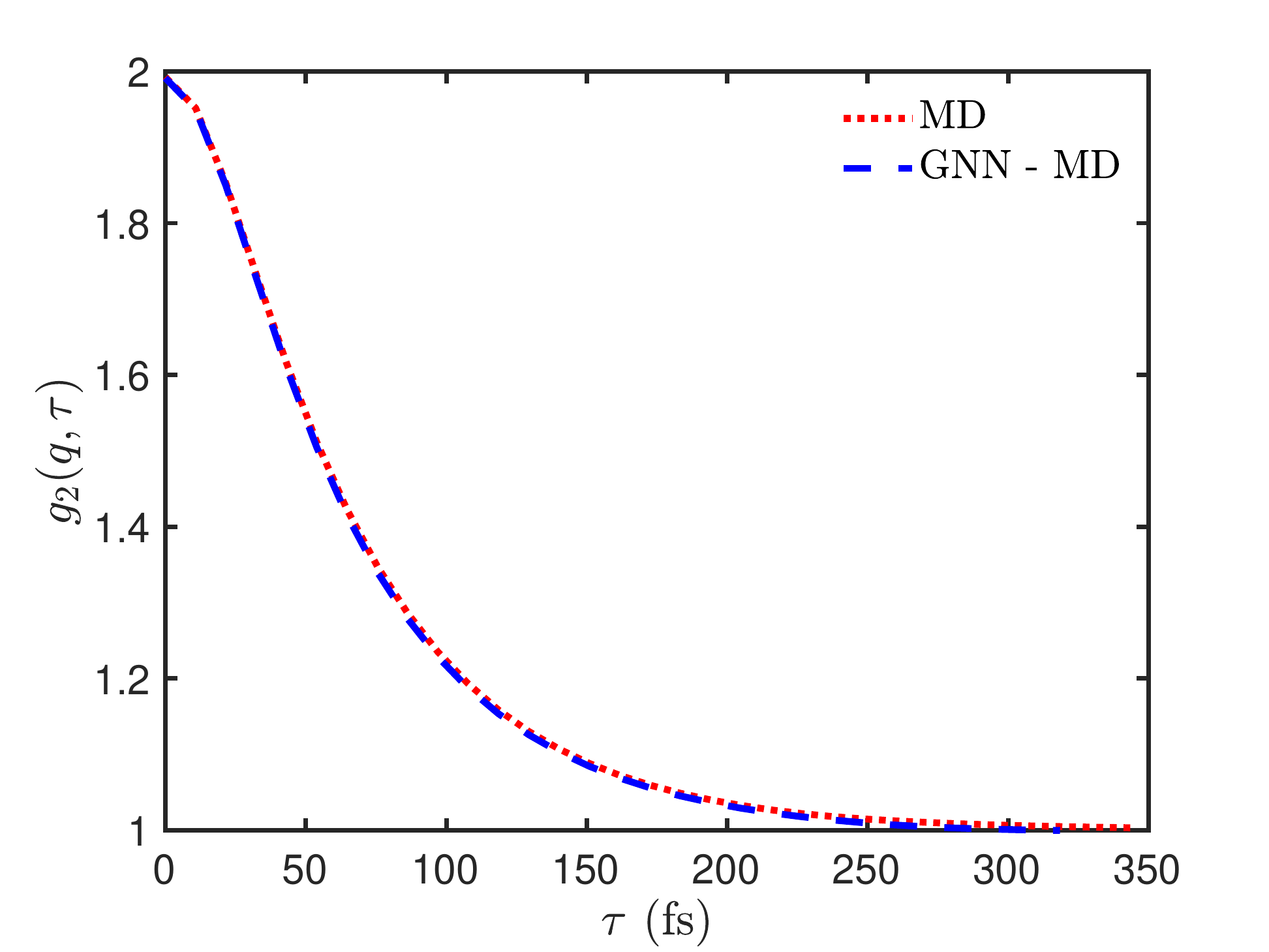}
    \label{fig:g2_4000}}
     \caption{The $g_2(q,\tau)$ at $q = 1.844 \pm 0.029$ \AA$^{-1}$ over the first 3.5574 ps for the (a) 256 atoms and the (b) 4000 atoms MD and GNN-MD simulation at 100 K.  \label{fig:g2_all}}
\end{figure}

The MSD curves from the entire ($\sim$ 539 ps) trajectories are fitted to straight (dashed) lines to obtain the predicted self-diffusivity $D$ (see Fig.~\ref{fig:msd_all}). The agreement between MD and GNN-MD is within $10\%$ for the 256-atom system, and within $5\%$ for the 4000-atom system, at the temperature of 100~K. The level of agreement for the 4000-atom system is maintained at all temperatures from 95~K to 110~K, and appears to improve with increasing temperature, as shown in Table~\ref{tab:msd_diff}.
\begin{table}[H]
    \centering
    \begin{tabular}{c|c|c|c}
    \hline
    \multirow{2}{*}{Temperature} & \multicolumn{3}{c}{$D$ ($\mu$m$^2$/s)} \\
    \cline{2-4}
     & MD & GNN-MD &Relative Error ($\%$) \\
     \hline
    95 K & 3254.70 & 3096.14 & 4.87\\
    100 K  & 3673.08 & 3802.89& 3.53\\
    110 K & 5010.12& 4979.92&0.60\\
    \hline
    \end{tabular}
    \caption{The self-diffusivity calculated from the MSD for the 4000 atom system at different temperatures using the MD and GNN-MD simulation. The MSD for all temperatures can be found in Fig.~\ref{fig:app-msd_all}.}
    \label{tab:msd_diff}
\end{table}
In addition to the bulk diffusivity, we also test the GNN-based force field by carrying out the XPCS analysis. This allows us to quantify the density fluctuations arising from atomic motion at different length scales. 
Figure~\ref{fig:slices} shows two computed XPCS speckle patterns from snapshots of the MD and GNN-MD trajectories, respectively.  The speckle patterns are computed for all saved configurations ($\sim$ 107.8 fs apart) in the atomic trajectories so that the time correlation at each pixel can be obtained.
Figure~\ref{fig:g2_all} shows that the decay of the time correlation of the speckle averaged over all wave-vectors close to a given $q=\vert\bm{q}\vert$, agrees well between MD and GNN-MD, for both the 256-atom system and the 4000-atom system.
If the system dynamics is purely diffusive, then $g_2(q,\tau)$ should decay exponentially.
Figure~\ref{fig:g2_all} compares the time correlation for $q = 1.844 \pm 0.029$ \AA$^{-1}$ (probing the length scale corresponding to the first nearest neighbor).
We have also carried out the same computation for all wave-vector magnitudes from $q = 0.461 \pm 0.029$ \AA$^{-1}$ to $q = 1.844 \pm 0.029$ \AA$^{-1}$ and observed close agreement between MD and GNN-MD models.
In all cases, we see that the $g_2(q,t)$ decays as a single exponential after the initial sub-diffusive dynamics associated with caging effects~\cite{Mohanty2022}, as is expected for liquid Ar.
\begin{figure}[H]
    \centering
    \subfigure[]{\includegraphics[width=0.48\textwidth]{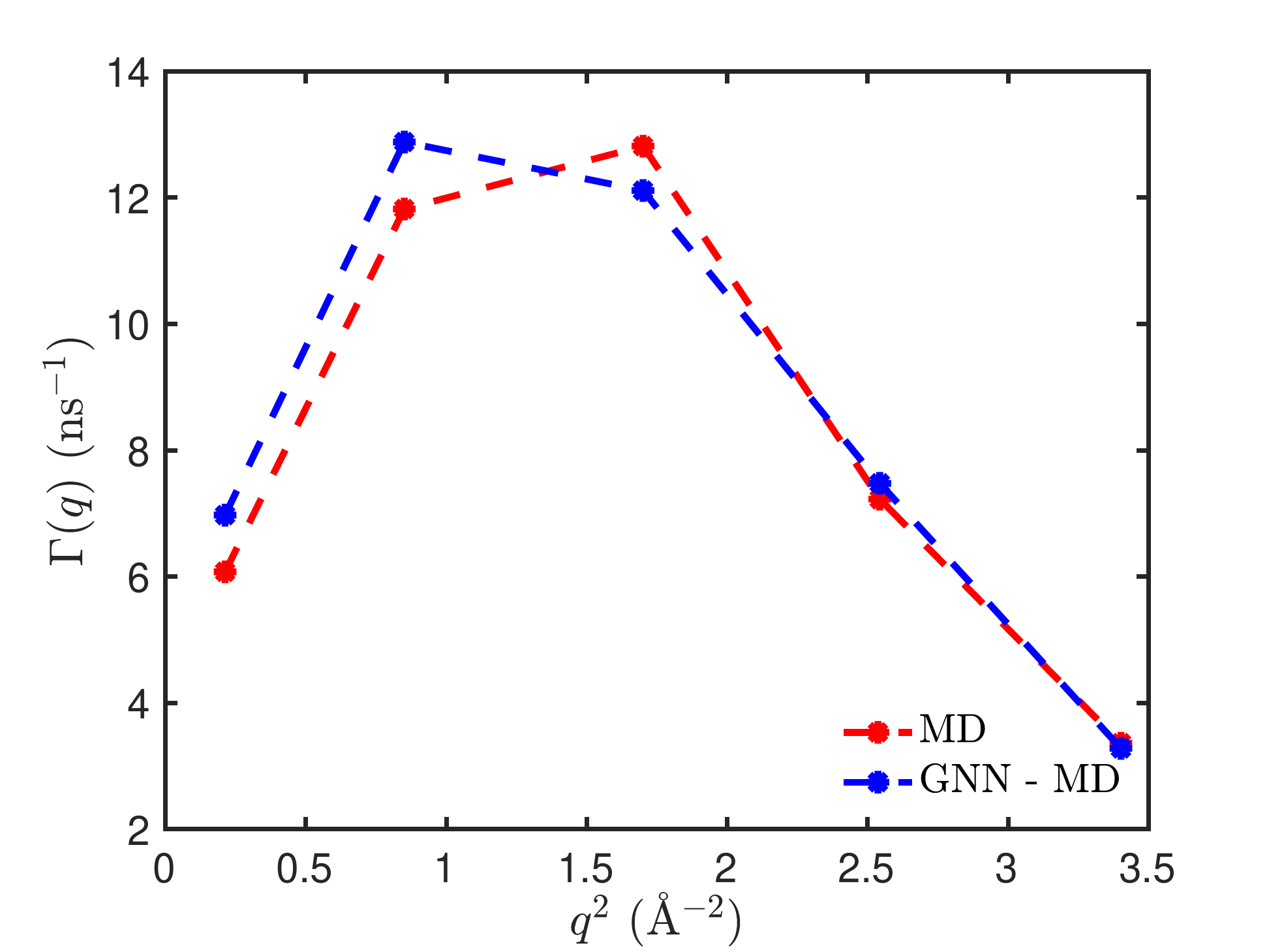}
    \label{fig:gamma_256}}
    \subfigure[]{\includegraphics[width=0.48\textwidth]{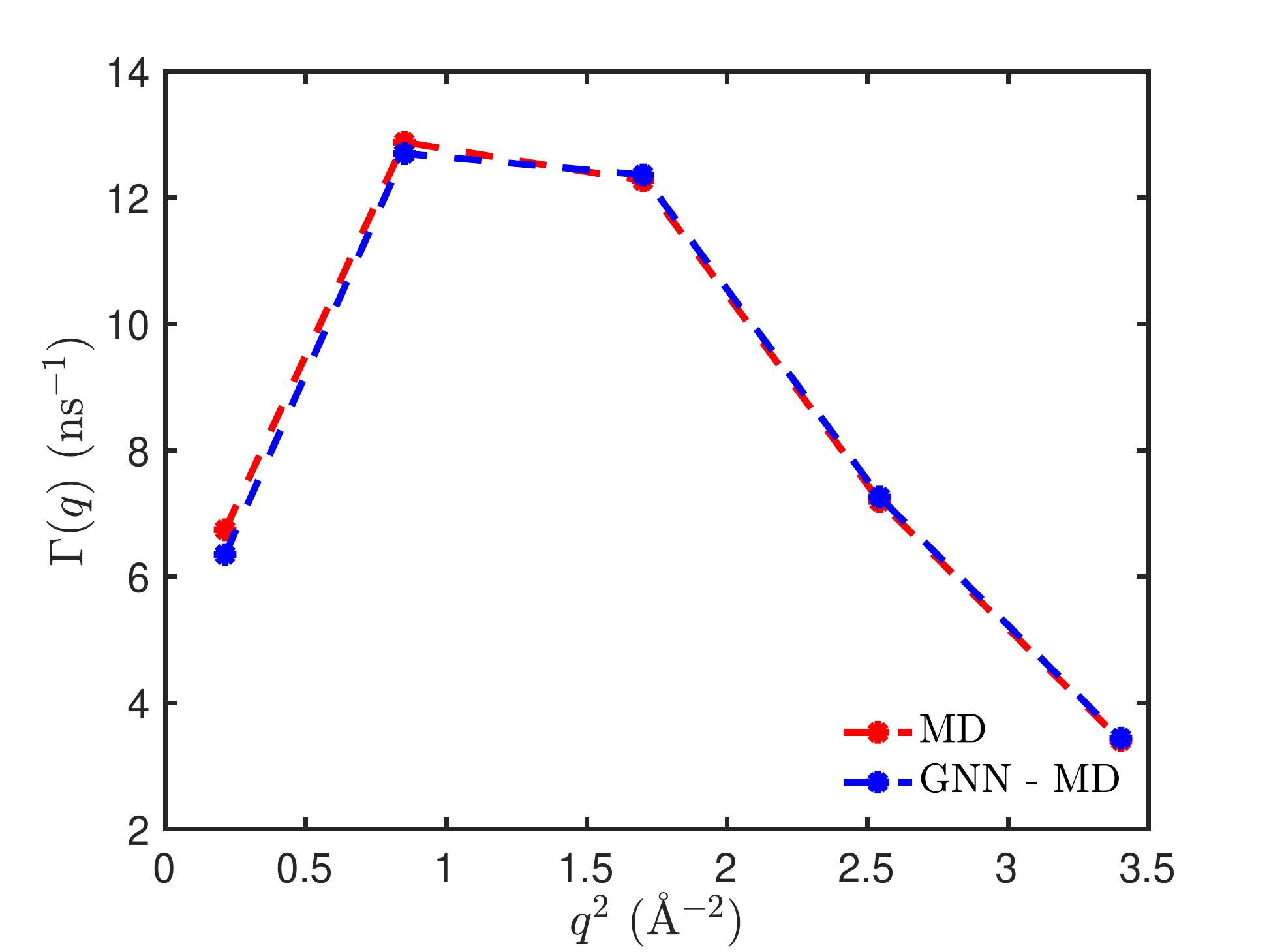}
    \label{fig:gamma_4000}}
    \caption{The $\Gamma(q)$ as a function of $q^2$ for the (a) 256 atoms and the (b) 4000 atoms MD and GNN-MD simulation at 100~K.  \label{fig:gamma_all}}
\end{figure}
From $g_2(q,\tau)$, we extract the rate of exponential decay $\Gamma(q)$ in the long time limit, which reveals the dispersion relation in the liquid~\cite{Mohanty2022}.
Figure~\ref{fig:gamma_all} shows that the $\Gamma(q)$ function obtained from MD and GNN-MD are in good agreement.
Interestingly, the agreement appears better in the $4000$-atom system than in the $256$-atom system on which the GNN-base force field is trained.
Additionally, we found that the dispersion relation from the GNN-MD simulation agrees well with MD simulation results for all temperatures of 95~K, 100~K, and 110~K (see Fig.~\ref{fig:app-gamma_all}).

\begin{figure}[H]
    \centering
    \subfigure[]{\includegraphics[width=0.48\textwidth]{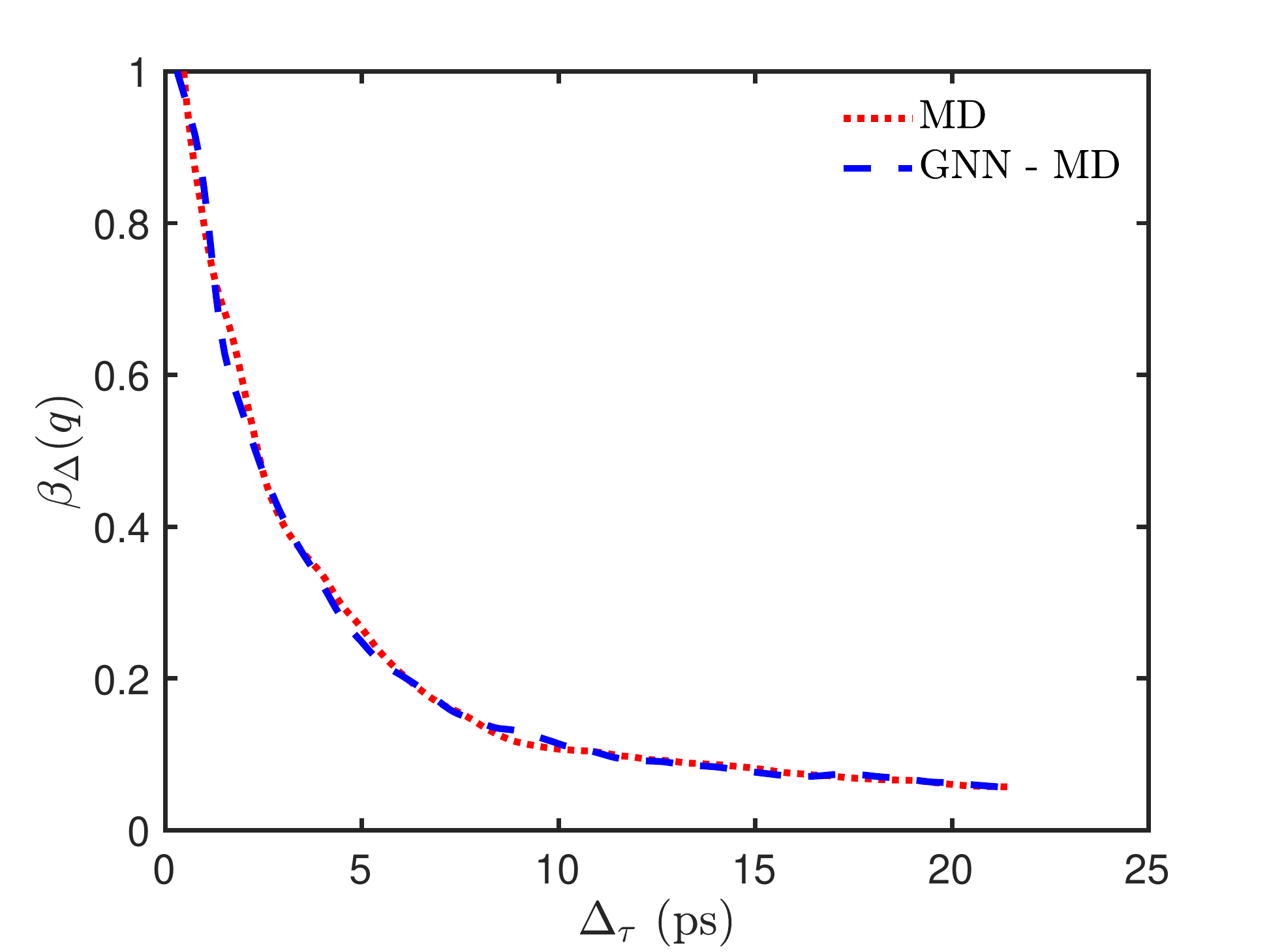}
    \label{fig:beta_256}}
    \subfigure[]{\includegraphics[width=0.48\textwidth]{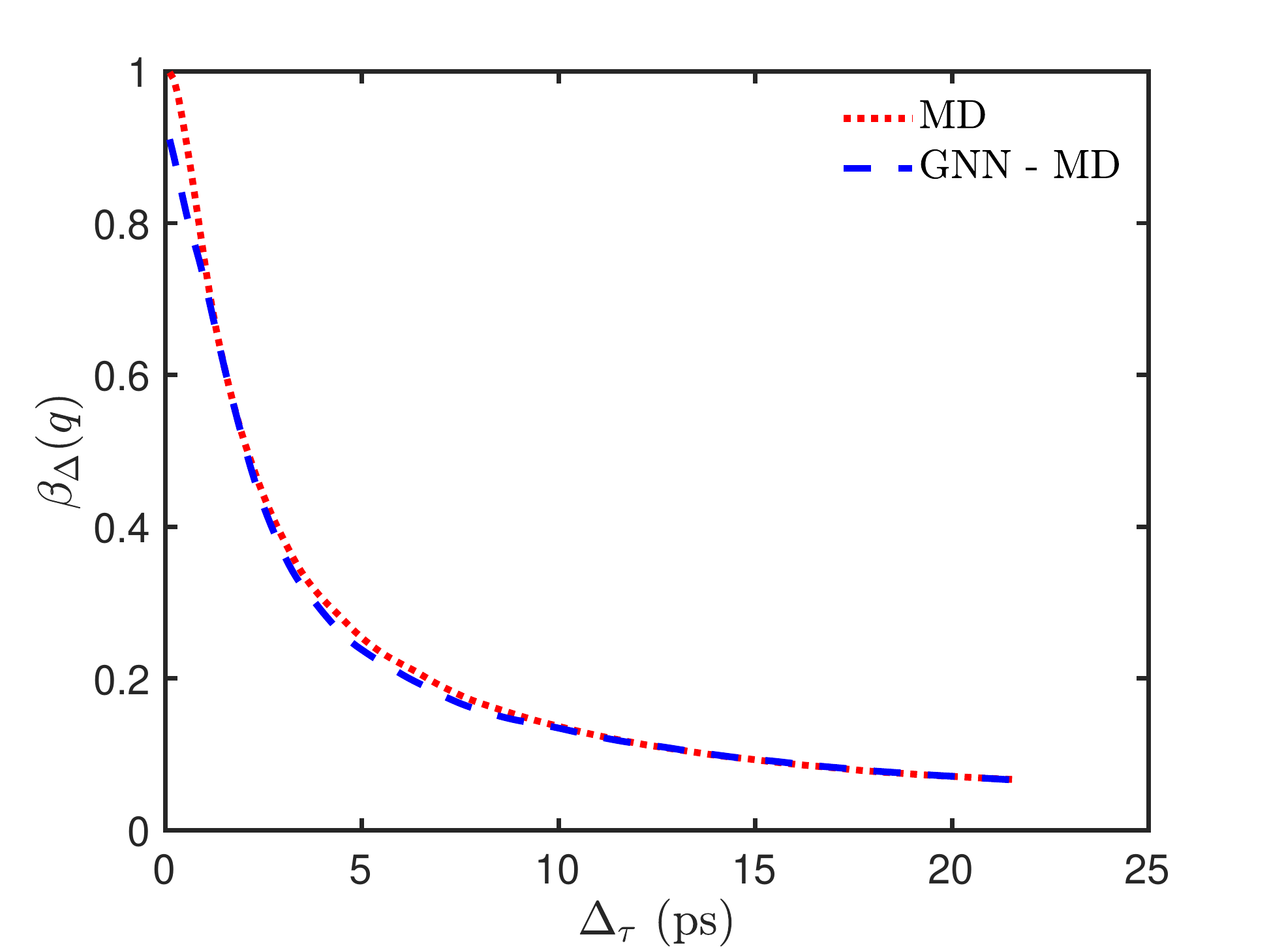}
    \label{fig:beta_4000}}
    \caption{The optical contrast $\beta_{\Delta}(q)$ as a function of X-ray pulse width ($\Delta_{\tau}$) for the (a) 256 atoms and the (b) 4000 atoms MD and GNN-MD simulation for $\lvert \bm{q} \rvert = 1.844 \pm 0.029$ \AA$^{-1}$ at 100 K.  \label{fig:beta_all}}
\end{figure}
In addition to the dynamics obtained from the XPCS speckle fluctuations, we also test the statistical distribution of the X-ray speckles by evaluating the optical contrast in the $q$-range $\lvert \bm{q} \rvert = 1.844 \pm 0.029$ \AA$^{-1}$ for varying pulse widths, $\Delta_{\tau}$, of the incident X-ray.
Figure~\ref{fig:beta_all} shows that the optical contrast obtained from MD and GNN-MD models are in close agreement. 
The optical contrast $\beta_{\Delta}(q)$ is expected to decrease with increasing pulse width, $\Delta_{\tau}$, where the time at which $\beta_{\Delta}(q)$ decreases to half of its maximum value gives an estimate of the correlation time.
Table~\ref{tab:corr_time} lists the correlation time obtained at three temperatures, where the estimates from MD and GNN-MD models agree well with each other (time resolution of 0.11 ps). 
\begin{table}[H]
    \centering
    \begin{tabular}{c|c|c|c}
    \hline
    \multirow{2}{*}{Temperature} & \multicolumn{3}{c}{Correlation time  (ps)} \\
    \cline{2-4}
     & MD & GNN-MD& Relative Error ($\%$) \\
     \hline
    95 K & 2.16 & 2.25 & $\sim$ 4.17 \\
    100 K  & 2.05 & 2.05 & $\sim$ 0.0\\
    110 K & 1.83& 1.83 & $\sim$ 0.0\\
    \hline
    \end{tabular}
    \caption{The correlation time (time taken for $\beta_{\Delta}(q)$ to drop to 0.5) for the 4000 atom system at different temperatures using the MD and GNN-MD simulation. The $\beta_{\Delta}(q)$ as a function of $\Delta_{\tau}$ for all temperatures can be found in Fig.~\ref{fig:app-beta_all}.}
    \label{tab:corr_time}
\end{table}
\subsection{Solid phase and melting} \label{sec:subsec:sol_phase}
\begin{figure}[H]
    \centering
   {\includegraphics[width=\textwidth]{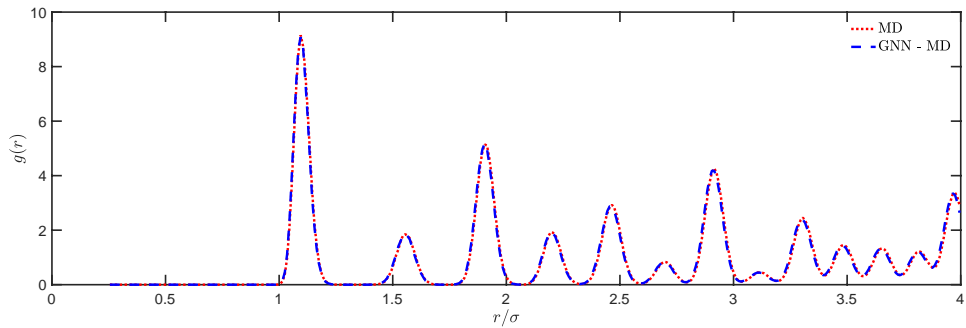}
   }
    \caption{The predicted pair distribution function $g(r)$ of the solid phase at 20~K using Model A, which is trained only using liquid configurations.  \label{fig:gr_nn}}
\end{figure}

In order to test the model performance in the solid phase, we performed MD and GNN-MD simulations of a face-centered cubic (FCC) lattice at 20~K, which is below the melting temperature.
The resulting pair distribution function $g(r)$ captures the amplitude of thermal vibration in the solid phase.
Figure~\ref{fig:gr_nn} shows that the GNN-MD captures the $g(r)$ of the solid phase accurately even for Model A, which is the GNN-based force field trained exclusively on liquid configurations.
Similar results for the predicted $g(r)$ of the solid phase are observed using Model B (not shown).

\begin{figure}[H]
    \centering
    \subfigure[]{\includegraphics[trim=2cm 4cm 0 4cm,clip, width=0.38\textwidth]{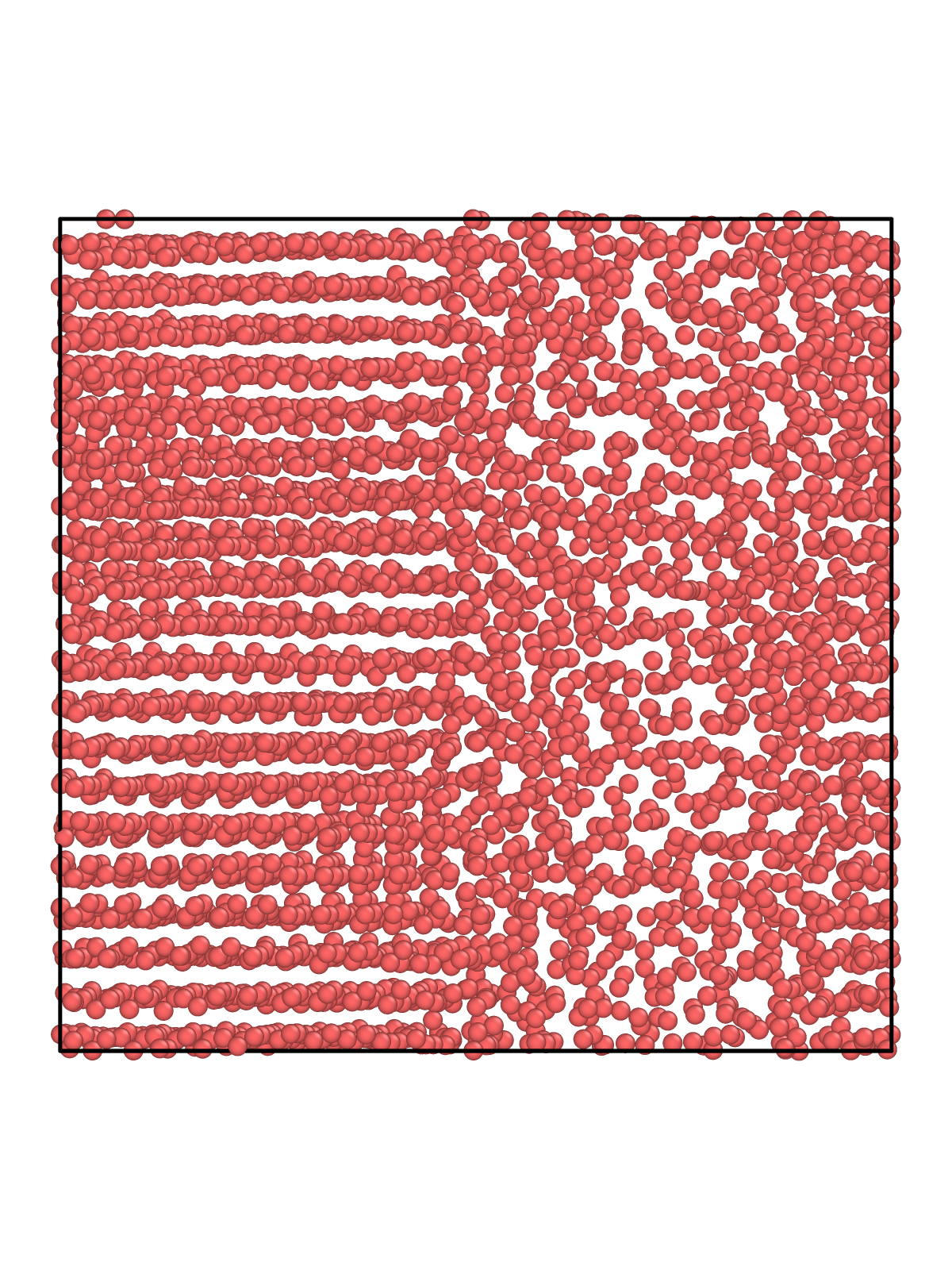}
    \label{fig:msd_melt}}
    \subfigure[]{\includegraphics[width=0.58\textwidth]{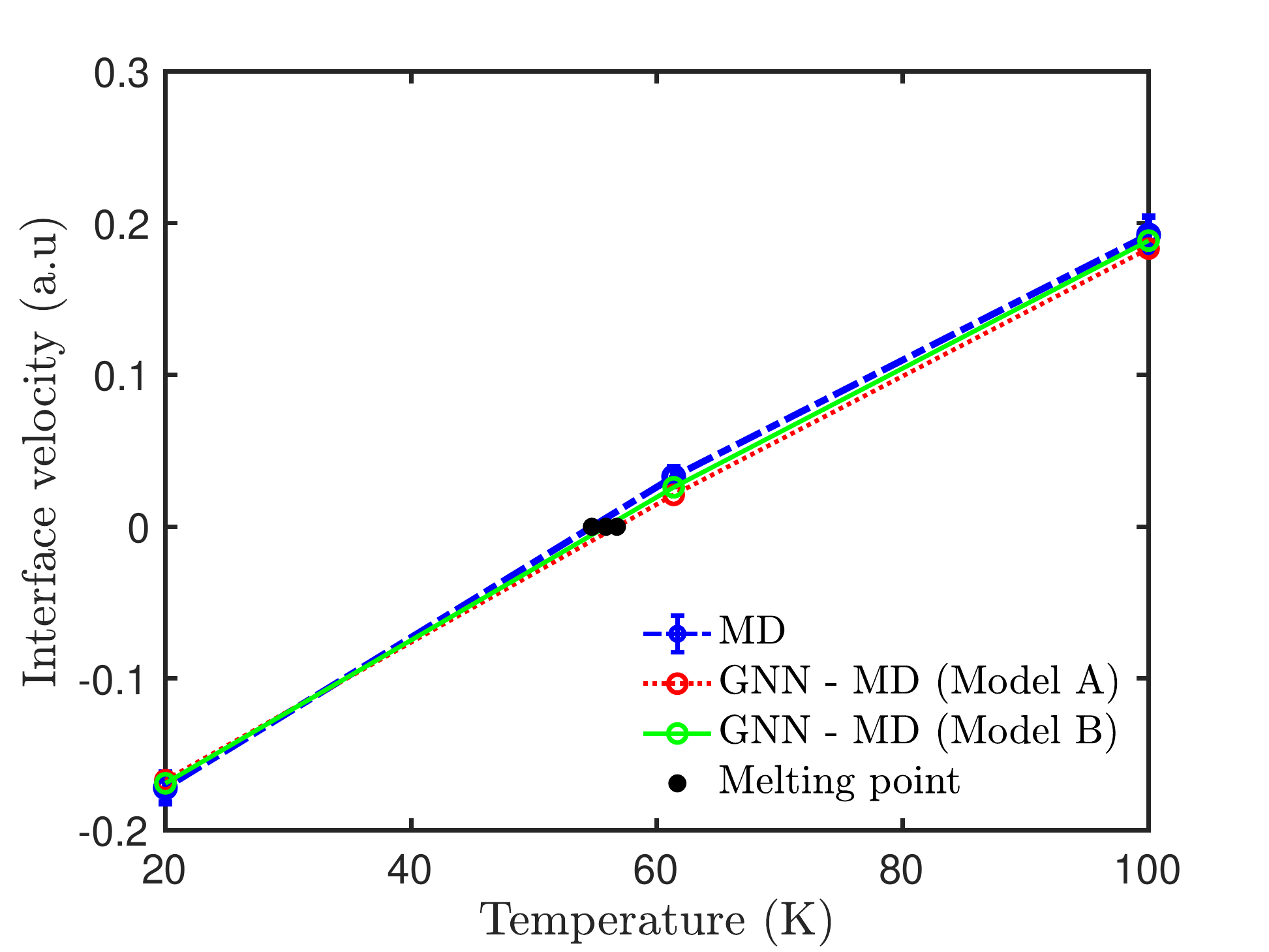}
     \label{fig:lf_melting}}
    \caption{(a) The solid-liquid interface used for the melting point estimation simulation and (b) the interface velocity at different temperatures for Model A and Model B.\label{fig:msd_misc}}
\end{figure}

To further test the capability of the model to capture solid phase configurations, we estimate the melting point of the system by using the solid-liquid interface method~\cite{zhu2021fully}. 
The solid half is initialized as an FCC lattice and is equilibrated at 20 K, whereas, the liquid half is equilibrated at 100 K to yield the initial configuration, as shown in Fig.~\ref{fig:msd_melt}.
The solid-liquid interface is maintained at 100 K, 61.36 K, and 20 K across three different simulations. The interface will move into the liquid or solid phase depending on whether the temperature is below or above the melting point.
To quantify these transitions we estimate the velocity of the solid-liquid interface. Our results show a linear dependence of the interface velocity with the simulation temperature, as shown in Fig.~\ref{fig:lf_melting}. Here the melting point corresponds to the temperature where the interface velocity is zero. We obtain the melting point from the regular MD simulation to be 55.2 K, whereas the melting point from the GNN-MD is: 56.4 K from Model A and 55.8 K from Model B. 
(The differences here are below the error bar $\sim 1.34$ K.)
This demonstrates that the GNN force field is able to capture the solid-liquid phase transition with sufficient accuracy. It is somewhat surprising that this level of agreement can be reached even for Model A, which is trained on liquid configurations only.

\begin{figure}[H]
    \centering
    \subfigure[]{\includegraphics[width=0.48\textwidth]{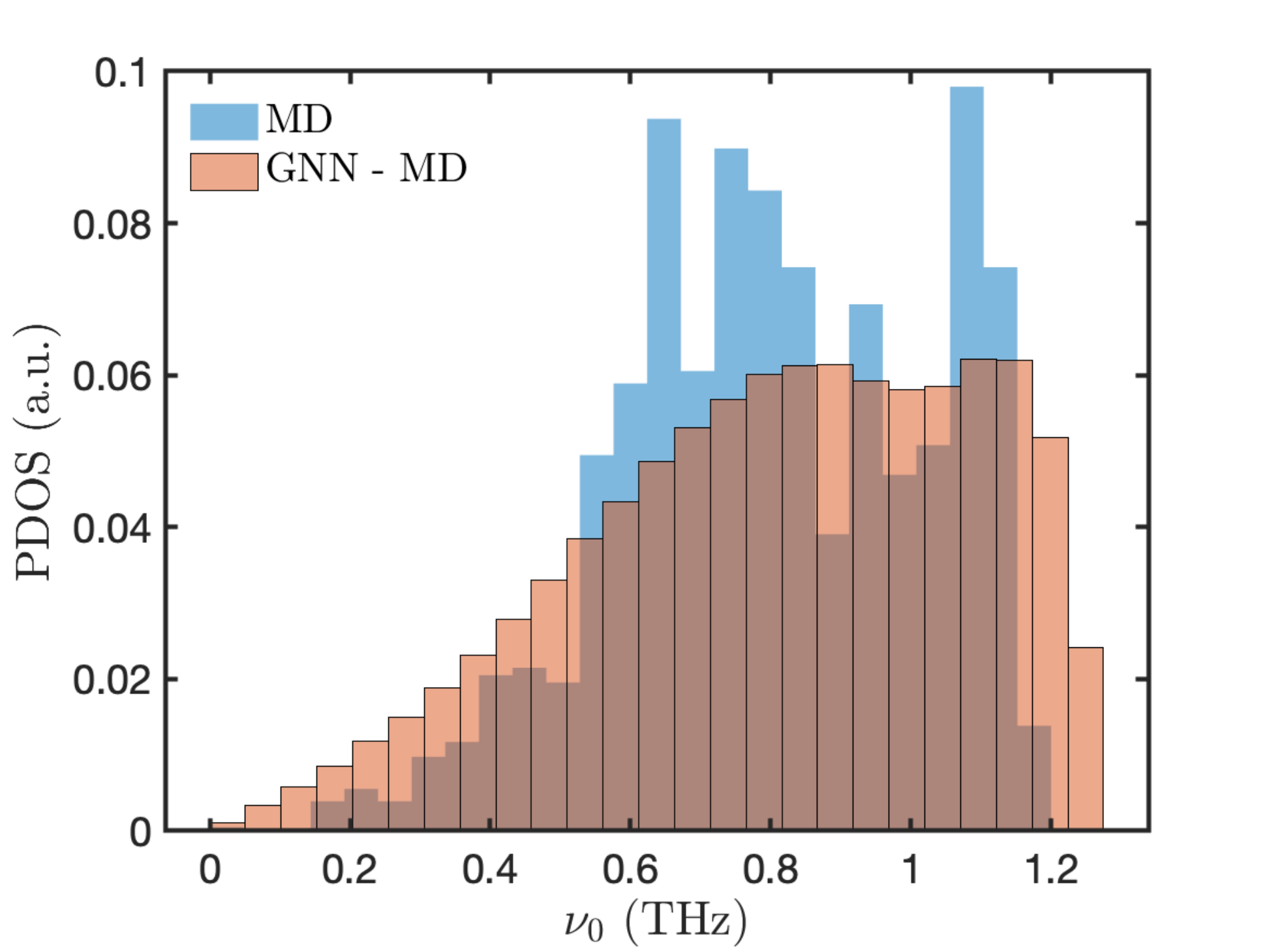}
   \label{fig:PDOS_liq}}
   \subfigure[]{\includegraphics[width=0.48\textwidth]{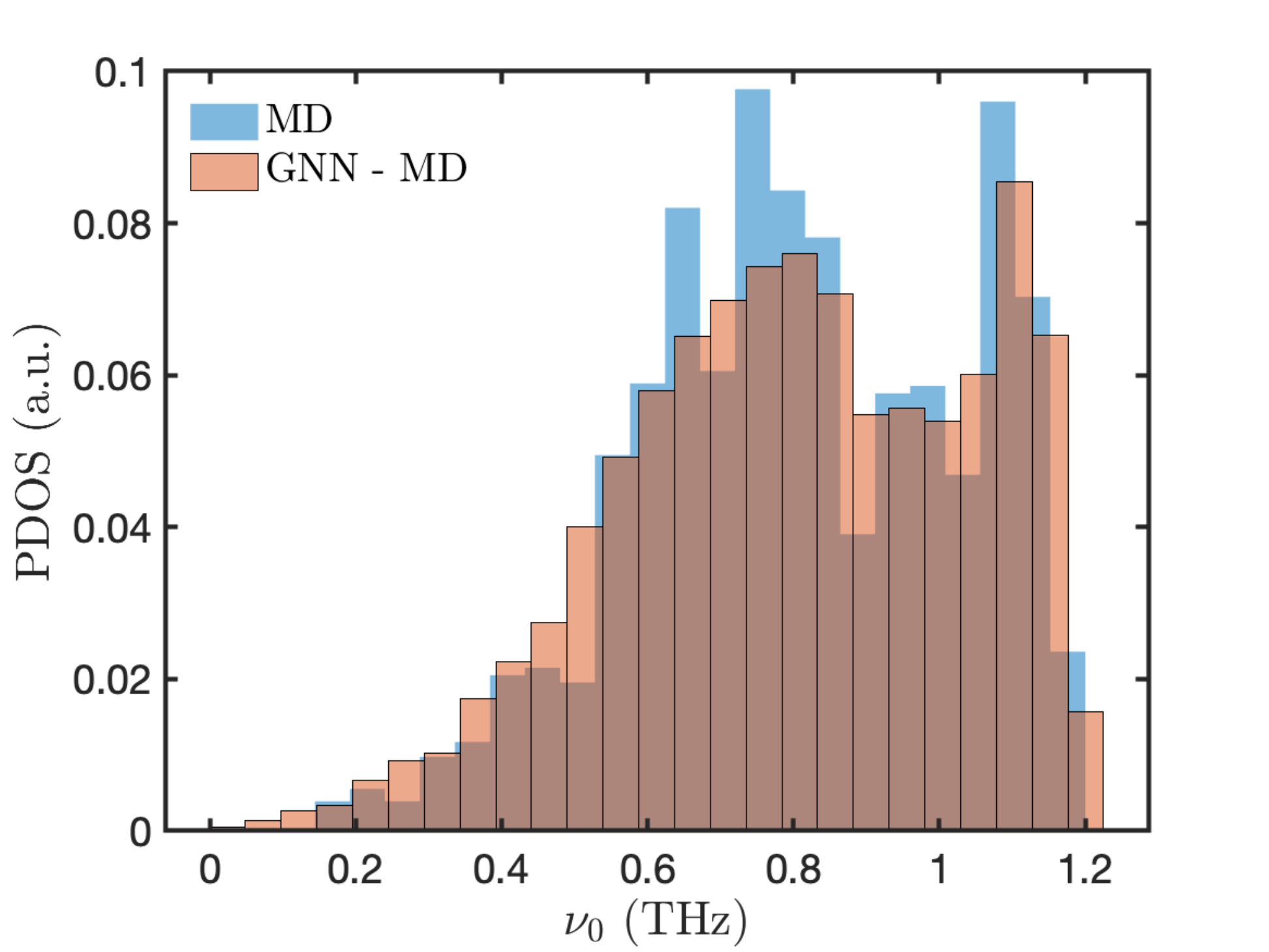}
   \label{fig:PDOS_both}}
    \caption{The phonon density of states obtained from Lennard-Jones force field computation and the GNN force field trained on: (a) exclusively liquid configurations (Model A), and (b) solid and liquid configurations (Model B). \label{fig:PDOS}}
\end{figure}

To see whether Model A can capture all the properties of the solid phase without being trained on any solid configurations, we performed further tests.
First, we examine the distribution of the eigenfrequencies of a perfect FCC crystal.
%, $\mathcal{D}$, which is 
The eigenfrequencies can be obtained by first diagonalizing the Hessian matrix, $\mathcal{H}$, and then dividing the eigenvalues by the atomic mass $m$ and taking the square root.
%divided by the mass of the particle, $m$. 
%
The distribution of the eigenfrequencies provides an estimate of the phonon density of states (PDOS) in the crystal, as shown in Fig~\ref{fig:PDOS}. For this analysis, we use a simulation cell with 2048 atoms, such that the $\mathcal{H}$ is defined over 6144 degrees of freedom. The details of the calculations can be found in \ref{subsec:PDOS}.
The Hessian matrix computed using the Lennard-Jones potential ($\mathcal{H}_{\rm LJ}$) is perfectly symmetric whereas the one computed from the GNN force field ($\tilde{\mathcal{H}}_{\rm GNN}$) is only approximately symmetrical. Here we use the $^\sim$ sign to indicate that the matrix is not exactly symmetrical.  Before computing the eigenfrequencies we need to symmetrize the Hessian matrix by $\mathcal{H}_{\rm GNN}^{\rm s}=\frac{1}{2}\left(\tilde{\mathcal{H}}_{\rm GNN} + \tilde{\mathcal{H}}_{\rm GNN}^{\rm T}\right)$.
Figure~\ref{fig:PDOS_liq} shows that the PDOS computed from GNN Model A (trained on liquid configurations only) exhibits an appreciable difference from the reference LJ model.
The agreement with the reference LJ model becomes significantly improved for GNN Model B (trained on both liquid and solid configurations), as shown in Fig.~\ref{fig:PDOS_both}.

The lack of perfect symmetry in the Hessian matrix is a type of error in the GNN-based force field considered here, in which the atomic forces are not obtained from the partial derivatives of a potential energy function.
To quantify this error, we define a \emph{symmetricity} measure $\mathcal{S}$ for any square matrix (see \ref{subsec:PDOS}) such that $\mathcal{S} = 1$ if the matrix is perfectly symmetric.

\begin{table}[H]
    \centering
    \begin{tabular}{c|c|c}
    \hline
    \multirow{2}{*}{Configuration} & \multicolumn{2}{c}{Training Data} \\
    \cline{2-3}
     & Liquid (Model A) & Solid and Liquid (Model B) \\
     \hline
    Solid & 0.785 & 0.949 \\
    Liquid  & 0.984& 0.984\\
    \hline
    \end{tabular}
    \caption{The symmetricity, $\mathcal{S}(\tilde{\mathcal{H}}_{\rm GNN})$, of the Hessian matrix of GNN Model A (trained on liquid configuration only) and Model B (trained on both liquid and solid configurations) when tested on a solid configuration (at $\sim 20$ K) and a liquid configuration (at $\sim 100$ K).}
    \label{tab:symm_values}
\end{table}

Table~\ref{tab:symm_values} shows the symmetricity of the Hessian matrix from Model A and Model B evaluated on a solid (perfect FCC crystal) and on a randomly chosen liquid configuration.
The symmetricity on the liquid configuration is 0.984 for both Model A and Model B, which is considered acceptable in this study.
The symmetricity on the solid configuration is only 0.785 for Model A, which is correlated with its poor prediction of the PDOS of the perfect crystal (Fig.~\ref{fig:PDOS_liq}).
On the other hand, after the solid configurations are included in the training set, Model B gives a symmetricity of 0.949 for the solid configuration, which is correlated with the improved prediction of the PDOS of the perfect crystal (Fig.~\ref{fig:PDOS_both}).
Therefore, our results demonstrate the need for a comprehensive set of tests to evaluate the transferability of machine-learned force fields, and the importance of including both liquid and solid configurations in the training dataset.

\section{Conclusions} \label{sec:conclusion}
This paper discusses the comprehensive lists of tests needed to check the transferability of a neural network force field. We carry out our benchmarking tests on the GNN-based force field presented in \cite{li2022graph}. Our analysis is based on a reference system of liquid Argon by modeling it as a Lennard-Jones fluid. The typical tests for a force field include validating the radial distribution function and mean-squared displacement. However, in this paper, we present additional tests to validate the dynamics from the GNN force field by carrying out the computational XPCS analysis. These tests are carried out at different system sizes and temperatures to further establish the transferability of the GNN force field.

In addition to testing the liquid behavior, we present a few solid-phase tests such as melting point estimation, which are not usually validated for machine-learned force fields. Notably, most solid and liquid phase tests were passed by the force field trained on just liquid configurations, however, estimating the phonon density of states required solid configurations in the training dataset. This form of dataset engineering is necessary for the force field to be able to capture both solid and liquid behavior.
In summary, these benchmarking tests provide a more comprehensive check on the transferability of the GNN-based force fields. 

On having established the transferability of the GNN force field, efforts can now be made to train the model from AIMD calculations for systems that are not well modeled by traditional MD force fields. This resulting GNN-AIMD promises the force field accuracy of AIMD and can be used for larger atomic systems which is not possible using traditional AIMD (due to computational expense) and traditional MD (due to the inaccuracy in the force field). 
\section*{Conflict of Interest}
The authors declare no competing interests.
\section*{Data Access}
All data is available in the main text and the Supplementary appendices. Further information about the computation can be obtained on request from the corresponding author. The library for the testbed of machine-learned force fields can be found at \href{https://gitlab.com/micronano_public/tb-mlff}{TB-MLFF}. The code on the XPCS and XSVS analysis is obtained from the \href{https://gitlab.com/micronano_public/c-xpcs}{C-XPCS} library. 
\section*{Acknowledgements}
S.M. and W.C. acknowledge support from the Precourt Pioneering Project of Stanford University. S.Y. was supported by Korea Institute for Advancement of Technology (KIAT) grant funded by the Korea Government (MOTIE) (P0017304, Human Resource Development Program for Industrial Innovation). K.K. acknowledges support from the National Research Foundation of Korea (NRF) grant funded by the Korean government (MSIT) (NRF-
2022R1A2C2011266).

\appendix
\newpage
\section{Analysis Details} \label{app:model}
\subsection{Mean-squared Displacement}
Mean squared displacement (MSD) is a metric used in statistical mechanics to measure the diffusive motion of the material system being simulated. The MSD is the most commonly used metric to measure the spatial deviation as a function of time. The MSD is an ensemble average and is computed using the position ($\bm{r} \in \mathcal{R}^d$ where $d$ is the dimensionality of the position),
\begin{equation}
    {\rm MSD}(\tau) = \langle (\bm{r}(\tau)-\bm{r}(0))^2 \rangle = \frac{1}{N}\sum_{i=1}^N (\bm{r}^i(\tau)-\bm{r}^i(0))^2.
\end{equation}
For a purely diffusive motion, such as the dynamics of Brownian motion, the MSD is related to the diffusivity, $D$, by,
\begin{equation}
    {\rm MSD}(\tau) = 2dD\tau.
\end{equation}
For the liquid simulations, the MSD varies linearly with time after the initial slow dynamics (attributed to the sub-diffusive motion brought about by the caging effect). As a consequence, the MSD is analyzed for $t>1$ ps.

\subsection{Basics of structural analysis} \label{subsec:RDF}
An important structural measure of a collection of atoms is the pair distribution function, $g(\bm{r},t)$, defined by,
\begin{equation}
    g(\bm{r},t)=\frac{1}{N\rho_{0}} \sum_{i} \sum_{\substack{j\\ j\neq i}} \delta (\bm{r}-(\bm{r}_{i}(t)-\bm{r}_{j}(t))),
\end{equation}
where $\rho_{0}$ is the bulk atomic density of the sample, $N$ is the total number atoms, and $\bm{r}_i$ (and $\bm{r}_j$) is the atomic position in the simulation box. 
Although the convention is to exclude the correlation between the atom and itself (i.e. $i=j$) from the definition of $g(\bm{r},t)$, in the following it is more convenient to introduce an alternative definition, $\tilde{g}(\bm{r},t)$, where such a constraint is removed.
\begin{equation}
    \tilde{g}(\bm{r},t)=\frac{1}{N\rho_{0}} \sum_{i} \sum_{j} \delta (\bm{r}-(\bm{r}_{i}(t)-\bm{r}_{j}(t)))
    = g(\bm{r},t)+\frac{1}{\rho_0}\,\delta(\bm{r})
\end{equation}
When all atoms are of the same type, the Fourier transform of $\rho_0\,\tilde{g}(\bm{r},t)$ is the structure factor, $\tilde{S}(\bm{q},t)$,
\begin{equation}
  \label{eq:sq_def_pure}
    \tilde{S}(\bm{q},t)=\frac{1}{N} \sum_{i}\sum_{j} 
    {\rm e}^{-\ui\bm{q}\cdot[\bm{r}_{i}(t)-\bm{r}_{j}(t)]}
    ,
\end{equation}
where $\bm{q}$ is the scattered wave-vector.
When the sample contains atoms of different types, the definition of the structure factor is generalized to the following
\begin{equation}
 \label{eq:Sq_def}
    S(\bm{q},t)=\frac{1}{\sum_{j}f_{j}(\bm{q})^{2}}\sum_{i}\sum_{j} f_{i}(\bm{q})f_{j}(\bm{q})\, {\rm e}^{-\ui\bm{q}\cdot[\bm{r}_{i}(t)-\bm{r}_{j}(t)]}
    .
\end{equation}
Here $f_{i}(\bm{q})$ is the X-ray atomic form factor of atom $i$. The X-ray atomic form factor can be obtained by the Fourier transform of the electron density field for a given type of atom.  
For many elements, the atomic form factor can be well parameterized by a sum of Gaussians \cite{prince2004international}.

\subsection{Basics of XPCS and scattering statistics} \label{subsec:XPCS}
We then consider the time correlation function of the intensity of the XPCS speckles that we observe. The normalized auto-correlation function helps in studying the dynamics of the simulation. The auto-correlation is denoted by $g_{2}(\tau)$, since it is a second order correlation and is given by,
\begin{align}
    g_{2}(\bm{q},\tau)&=\langle I(\bm{q},t) I(\bm{q},t+\tau) \rangle/\langle I \rangle ^{2} \\
    I(\bm{q},\tau) &\propto S(\bm{q},\tau)
\end{align}

Another metric that helps us in studying the dynamic characteristics for the simulation is the intermediate scattering function, $f(\bm{q},\tau)$, described as,
\begin{equation}
  f(\bm{q},\tau)=\frac{1}{N} \left\langle \sum_{i \neq j} e^{(\ui\bm{q}(r_{i}(0)-r_{j}(\tau))} \right\rangle _t
\end{equation}

For a simple diffusive process following Brownian motion, the intermediate scattering function reduces to a single exponential, $F(\bm{q},\tau)=\frac{f(\bm{q},\tau)}{f(\bm{q},0)}=e^{-\Gamma \tau}$, where $\Gamma=D \bm{q}^{2}$ \cite{chapman2006femtosecond}.

Since the time auto-correlation ($ g_{2}(\bm{q},\tau)$) is related to the $f(\bm{q},\tau)$ by the Siegert relation \cite{jakeman1974photon}, it can also be represented as a single exponential under the assumption of Brownian dynamics,

\begin{align}
g_{2}(\bm{q},\tau) & =1+\beta (\bm{q}) \lvert F(\bm{q},\tau) \rvert^{2} \nonumber \\
    & =1+\beta (\bm{q})  \exp[-2\Gamma(\bm{q}) \tau] \\
     g_{2}(\bm{q},\tau)-1 & =G(\bm{q},\tau) \propto \exp[-2\Gamma(\bm{q}) \tau]
\end{align}
where $\Gamma(\bm{q})$ is the relaxation rate and the factor of 2 arises from the Siegert relation which relates the intensity auto-correlation function to the electric field correlation function \cite{jakeman1974photon}. 

Furthermore, the auto-correlation of the speckle at $\bm{q}$ is denoted by $g_{2}(\bm{q}, \tau)$ and it is non-dimensionalized by the square of its mean at $\tau=0$, as given below for infinitesimally short pulses,
\begin{equation}
  \label{eq:g2_def}
    g_{2}(\bm{q},\tau) =\frac{\langle I(\bm{q},t)\, I(\bm{q},t+\tau) \rangle_{t}}{\langle I (\bm{q},t) \rangle_{t}^{2} }
    = \frac{\langle S(\bm{q},t)\, S(\bm{q},t+\tau) \rangle_{t}}{\langle S (\bm{q},t) \rangle_{t}^{2} },
\end{equation}
where $\langle \cdot \rangle_t$ represents the time-averaged value of the enclosed entity.
Experimentally, the X-ray pulses always have a finite duration.  Hence the intensity should be replaced by the time-averaged X-ray intensity, $I_{\Delta}(\bm{q})$, over the exposure duration of $\Delta_{\tau}$,
\begin{equation}
  \label{eq:inte_Iq}
    I_{\Delta}(\bm{q}) = \int_{t}^{t+\Delta_{\tau}} I(\bm{q},t) \,\ud t .
\end{equation}
In this case,
\begin{equation}
\label{eq:g2_def_pulse}
    g_{2}(\bm{q},\tau) =\frac{\langle I_{\Delta}(\bm{q},t)\, I_{\Delta}(\bm{q},t+\tau) \rangle_{t}}{\langle I_{\Delta}(\bm{q},t) \rangle_{t}^{2} } .
\end{equation}
For a X-ray speckle pattern obtained from a single X-ray pulse,  the optical contrast $\beta({q})$ is defined by the variance of the intensity divided by the square of its mean \cite{Mohanty2022}.
While $\beta(q)$ is given by the scattering intensity distribution $I(\bm{q})$ from an infinitesimally short pulse, we denote the optical contrast from an X-ray pulse of finite duration ${\Delta}_t$ as $\beta_{\Delta}(q)$ where,
\begin{equation}
\label{eq:bq_Iq}
    \beta_{\Delta}(q)=\frac{\langle I_{\Delta}(\bm{q})^{2} \rangle_{q}-\langle I_{\Delta} (\bm{q})\rangle_{q} ^{2}}{\langle I_{\Delta} (\bm{q})\rangle_{q}^{2}}, 
\end{equation}
where $\langle \cdot \rangle_{q}$ represents the average over all detector pixels that satisfy $q-dq/2\le \lvert \bm{q} \rvert < q+dq/2$, for a small $dq$.
For an ergodic and isotropic system, the distribution of pixel intensity at around a particular $\bm{q}$ is the same in $\bm{q}$-space and $t$, so that 
\begin{equation}
\label{eq:bq_bt}
 \langle I_{\Delta}(\bm{q}) \rangle_{q} \approx \langle I_{\Delta}(\bm{q}) \rangle_{t} , \quad \langle I_{\Delta}(\bm{q})^{2} \rangle_{q} \approx \langle I_{\Delta}(\bm{q})^{2} \rangle_{t} .
\end{equation}
This means that we can also define an optical contrast $\beta_0(\bm{q})$ from the time variation of the intensity at a single $\bm{q}$,  
i.e,
\begin{equation}
\label{eq:bq_g2}
    \beta_{0}(\bm{q})=g_{2}(\bm{q},\tau=0)-1. 
\end{equation}
For an ergodic and isotropic system, $\beta_{\Delta}(q)$ and $\beta_0(\bm{q})$ should be equal. The detailed discussion on the XPCS relations and the statistics of X-ray speckles, the computational algorithm and its implementation can be found in \cite{Mohanty2022}.

\subsection{Hessian matrix and phonon density of states} \label{subsec:PDOS}
To estimate the phonon density of states (PDOS) we examine the eigenvalues of the Hessian matrix, $\mathcal{H}$, of our relaxed system. The $\mathcal{H}$ is defined as,
\begin{align}
    \mathcal{H}_{ij}&=\frac{\partial}{\partial x_j} \left(\frac{\partial U}{\partial x_i}\right), \\
    &= -\frac{\partial {f}_i}{\partial x_j},
\end{align}
where $U$ is the potential energy and ${f}_i$ is the atomic force component corresponding to the $i^{\rm th}$ degree of freedom ($i$ goes from 1 to $3N$).
We can approximate this derivative numerically by a central difference scheme, 
\begin{equation}
    \mathcal{H}_{ij} = \frac{{f}_i(x_j-\Delta x)-{f}_i(x_j+\Delta x)}{2 \Delta x},
\end{equation}
where $\Delta x = 3.405 \times 10^{-6}$ \AA ($10^{-6}$ in LJ unit). 
If $\lambda_n$ represents the eigenvalues of $\mathcal{H}$ then the eigenfrequencies $\nu_n$ can be obtained from
\begin{equation}
    \nu_n = \frac{1}{2\pi}\sqrt{\frac{\lambda_n}{m}}.
\end{equation}
The distribution of the $\nu_n$ gives us the PDOS in arbitrary units. 
The Hessian matrix $\mathcal{H}$ is expected to be symmetric for a conservative force field that is derived from an interatomic potential. Consistency or symmetricity requires,
\begin{equation}
    \frac{\partial {f}_i}{\partial x_j}=\frac{\partial {f}_j}{\partial x_i}.
\end{equation}
The Hessian matrix computed using the Lennard-Jones potential ($\mathcal{H}_{\rm LJ}$) is perfectly symmetric whereas the one computed from the GNN force field ($\tilde{\mathcal{H}}_{\rm GNN}$) is only nearly symmetric. We define a measure for symmetricity, $\mathcal{S}$, for a matrix $A$ as
\begin{equation}
    \mathcal{S}(A) = \frac{\vert\vert A_{\rm s} \vert\vert_2-\vert\vert A_{\rm a} \vert\vert_2}{\vert\vert A_{\rm s} \vert\vert_2+\vert\vert A_{\rm a} \vert\vert_2},
\end{equation}
where,
\begin{align}
    A_{\rm s} &= \frac{A+A^T}{2},\\
    A_{\rm a} &= \frac{A-A^T}{2}.
\end{align}
The symmetricity is such that $-1\leq \mathcal{S}(A)\leq 1$, where $\mathcal{S}(A)=1$ for a perfectly symmetric matrix and $\mathcal{S}(A)=-1$ for a perfectly anti-symmetric matrix. 
\newpage
\section{Analysis at different temperatures}
 In this appendix, we show the results corroborating our findings that have been presented in main text. We present simulation results corresponding to the 4000 atoms case which is run at 95 K, 100 K, and 110 K.
\begin{figure}[H]
    \centering
    \subfigure[]{\includegraphics[width=0.48\textwidth]{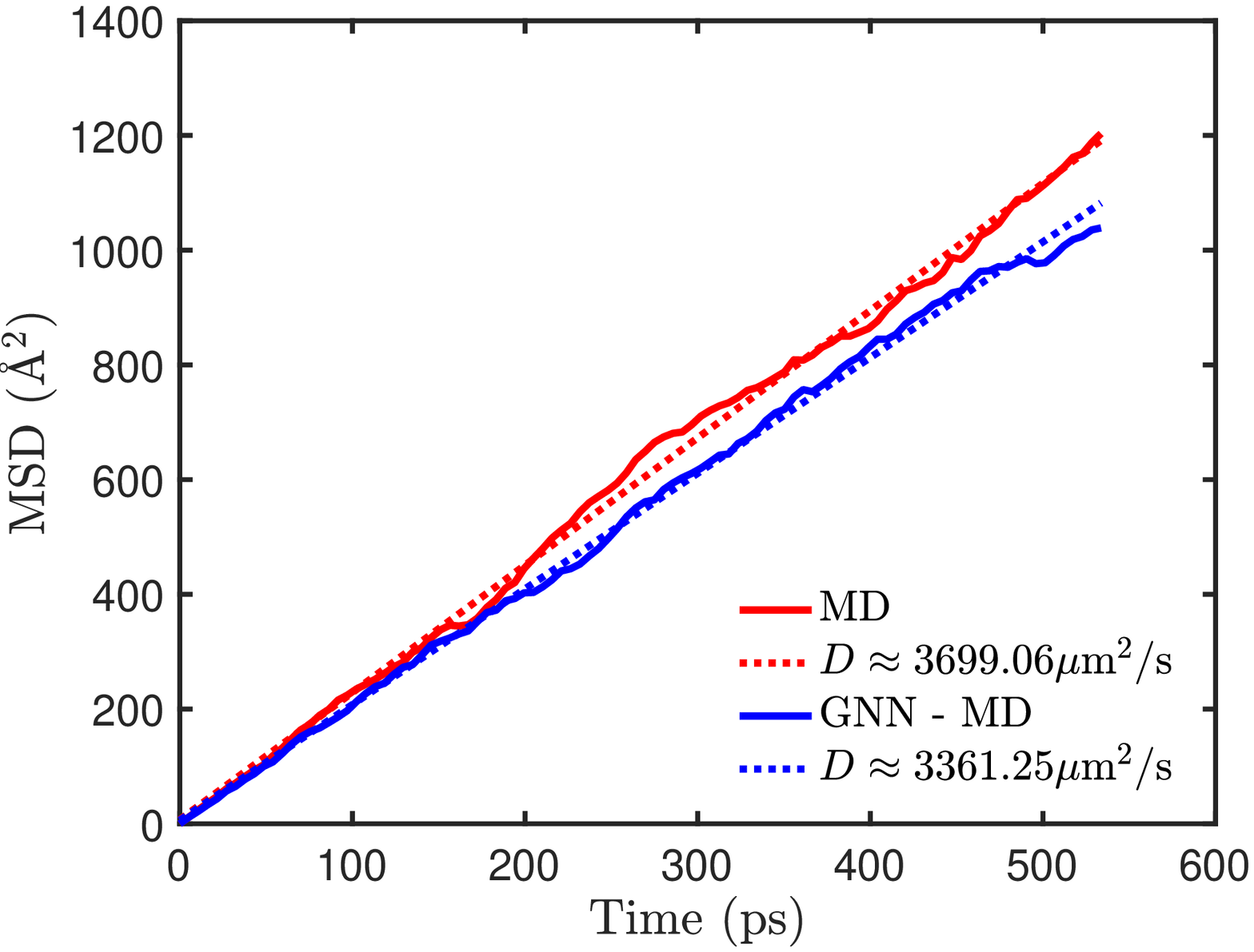}
    \label{fig:app-msd_256}}
    \subfigure[]{\includegraphics[width=0.48\textwidth]{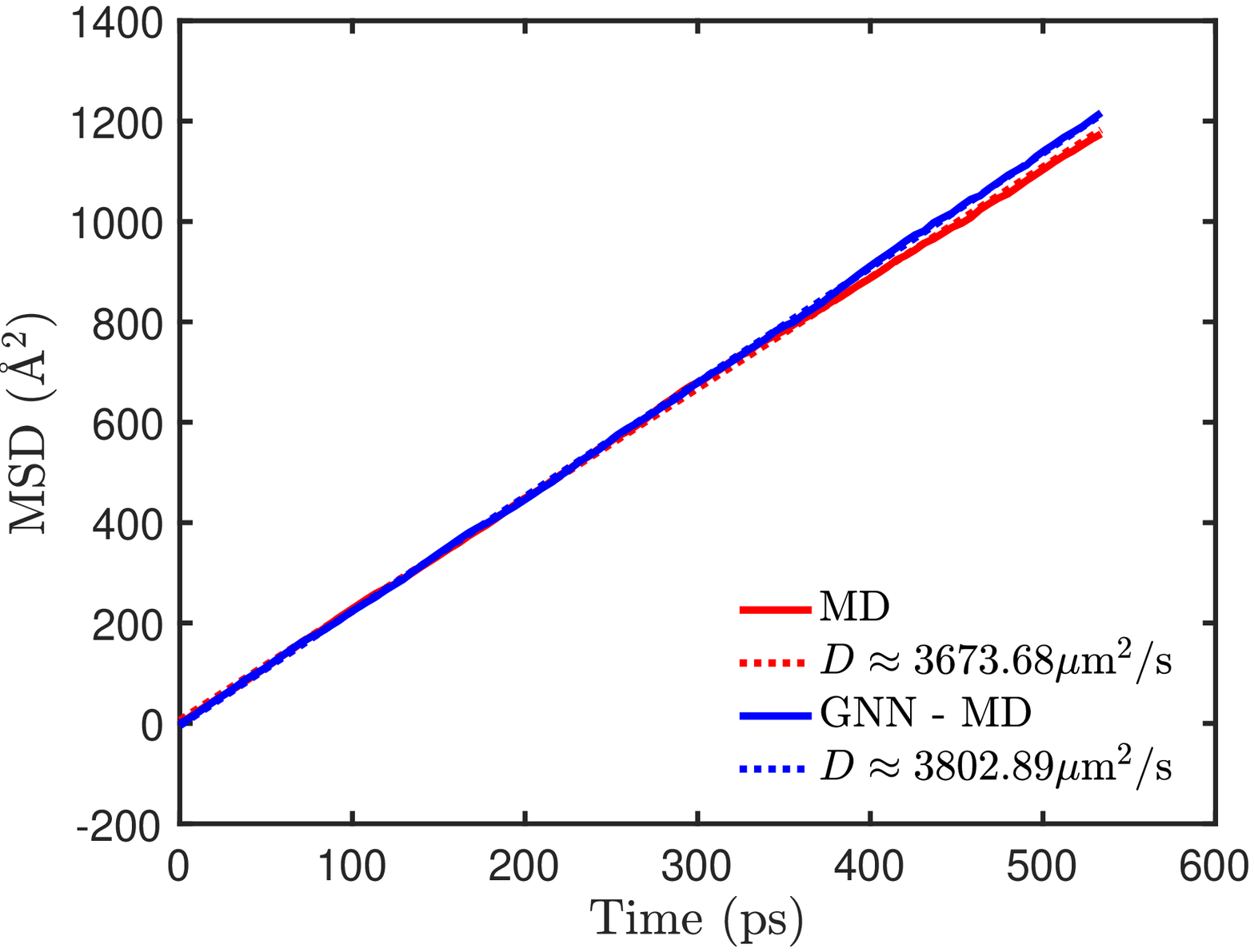}
    \label{fig:app-msd_4000}}
    \subfigure[]{\includegraphics[width=0.48\textwidth]{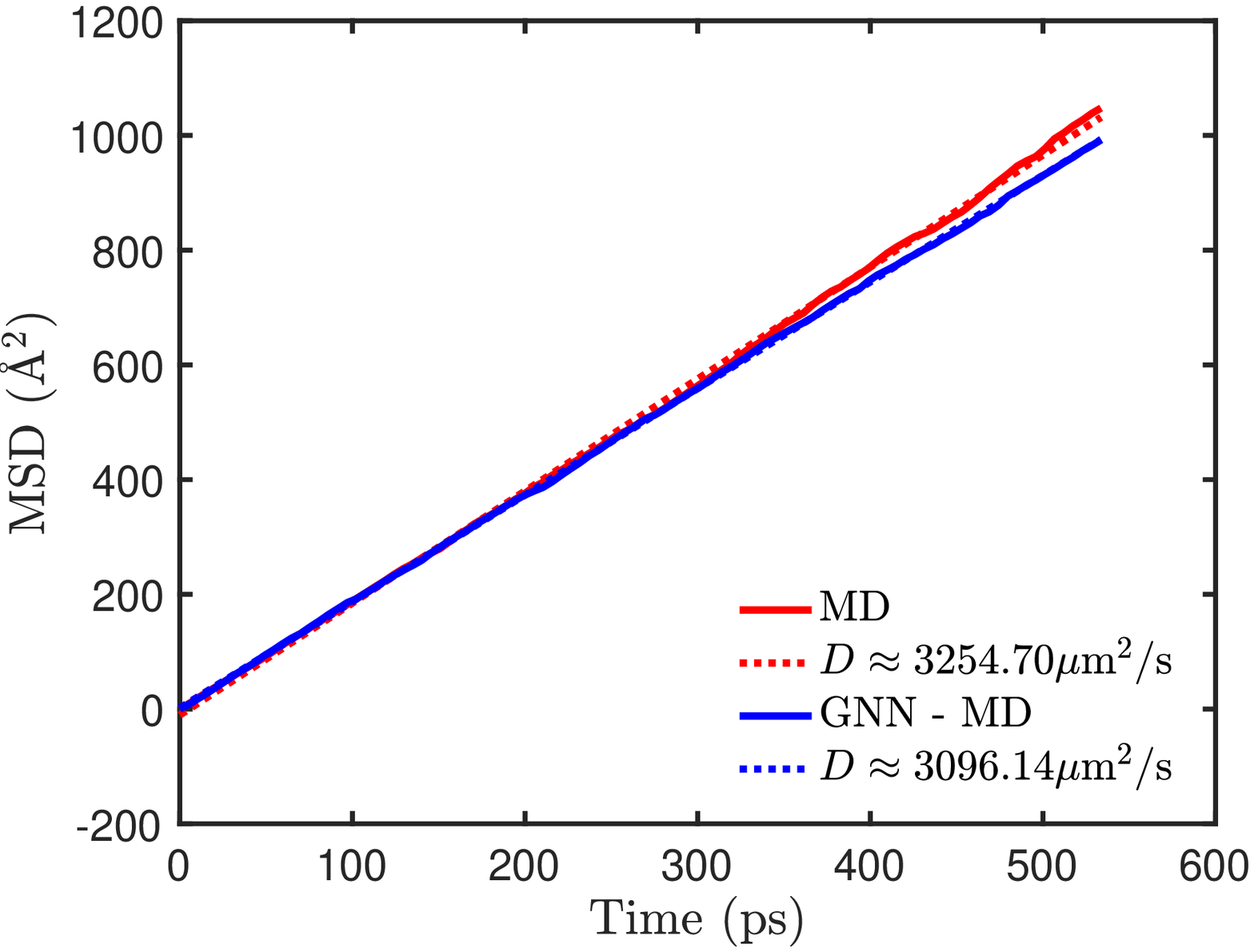}
    \label{fig:app-msd_95}}
    \subfigure[]{\includegraphics[width=0.48\textwidth]{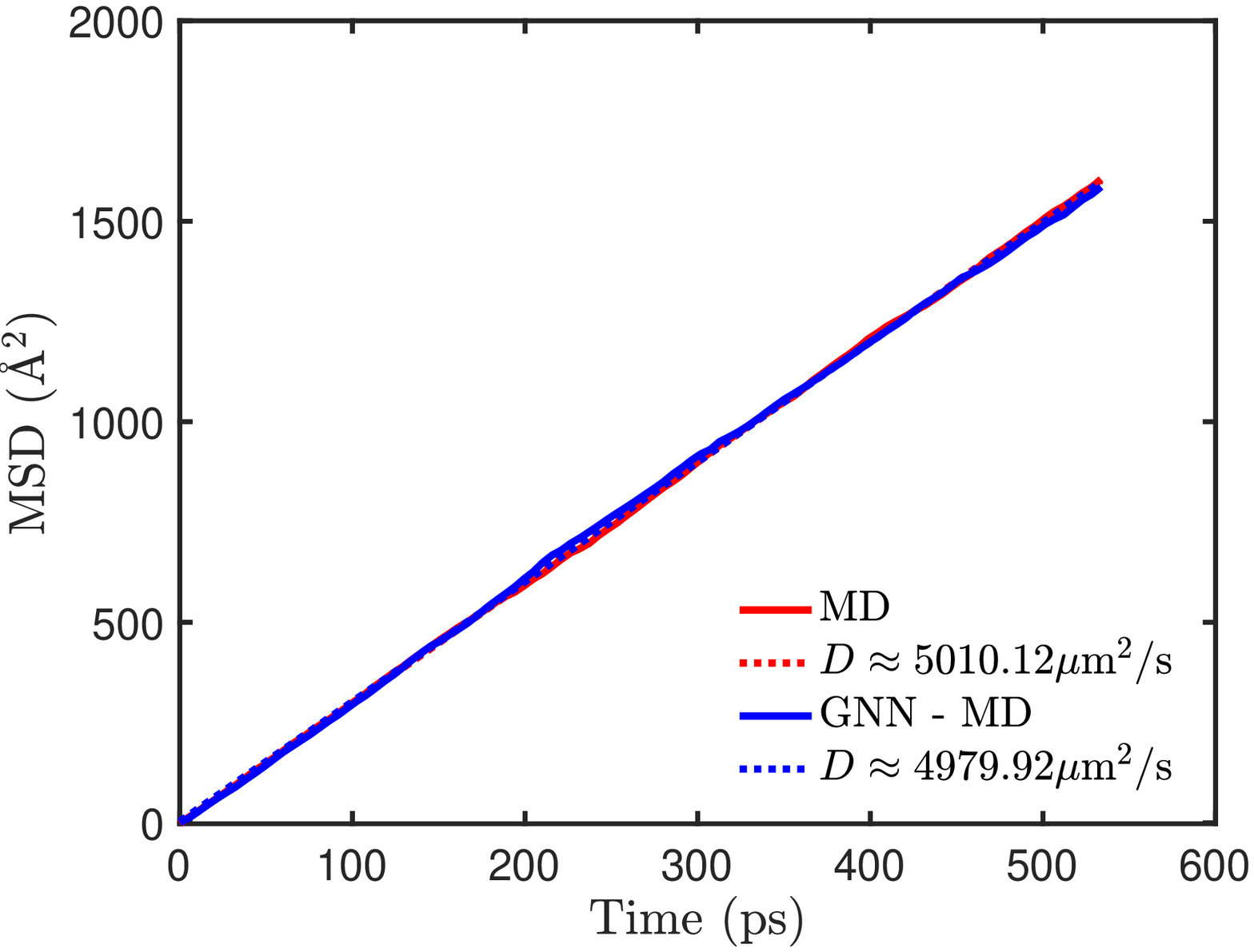}
    \label{fig:app-msd_110}}
    \caption{The mean-squared displacement (MSD) for the (a) 256 atoms at 100 K and the  4000 atoms at (b) 100 K, (c) 95 K, and (d) 110 K, MD and GNN-MD simulation over the 539 ps trajectory.  \label{fig:app-msd_all}}
\end{figure}
We see that the MSD is in close agreement between the MD and GNN-MD simulations, as shown in Fig.~\ref{fig:app-msd_all}, irrespective of the temperature at which the simulations are carried out.
\begin{figure}[H]
    \centering
    \subfigure[]{\includegraphics[width=0.48\textwidth]{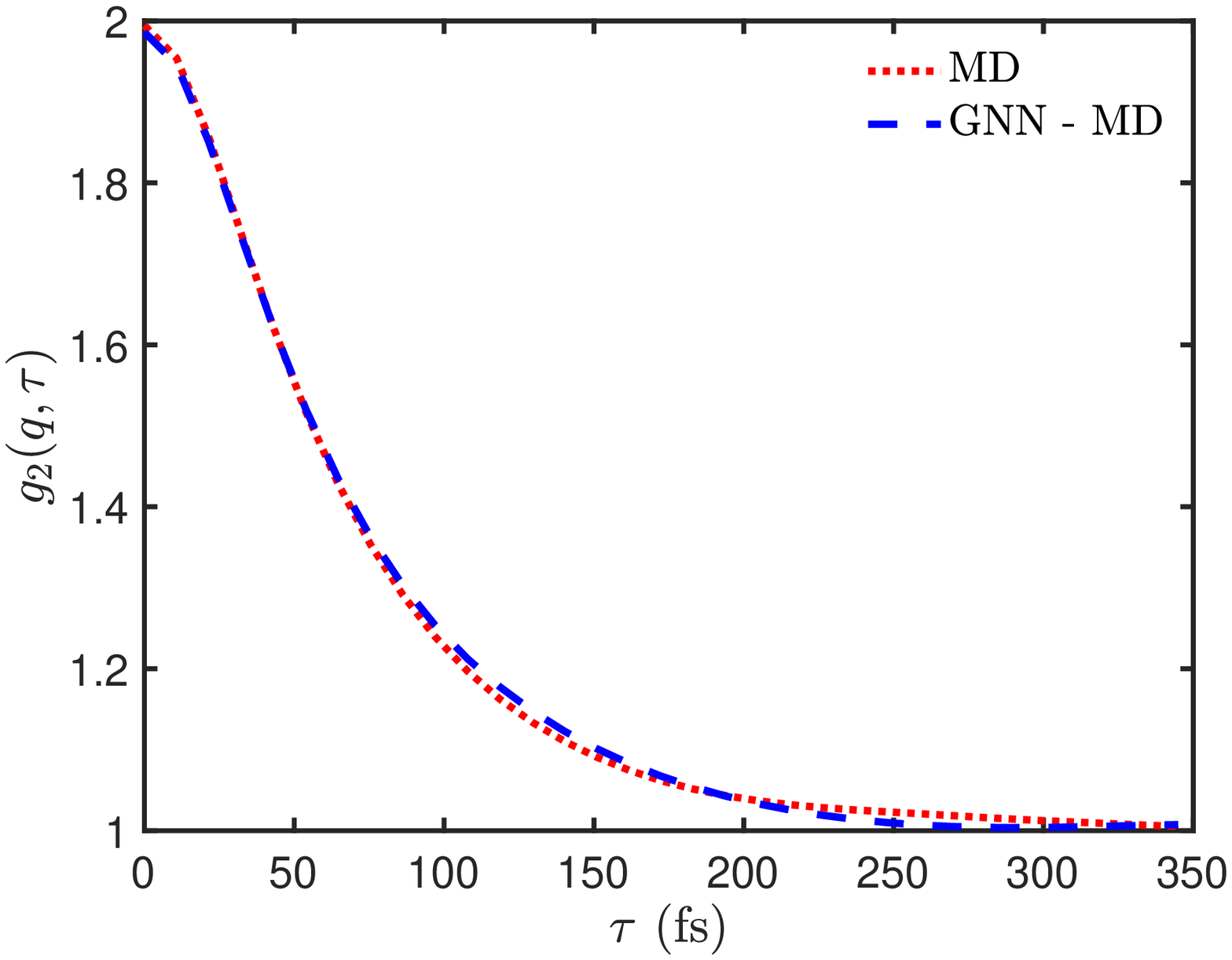}
    \label{fig:app-g2_256}}
    \subfigure[]{\includegraphics[width=0.48\textwidth]{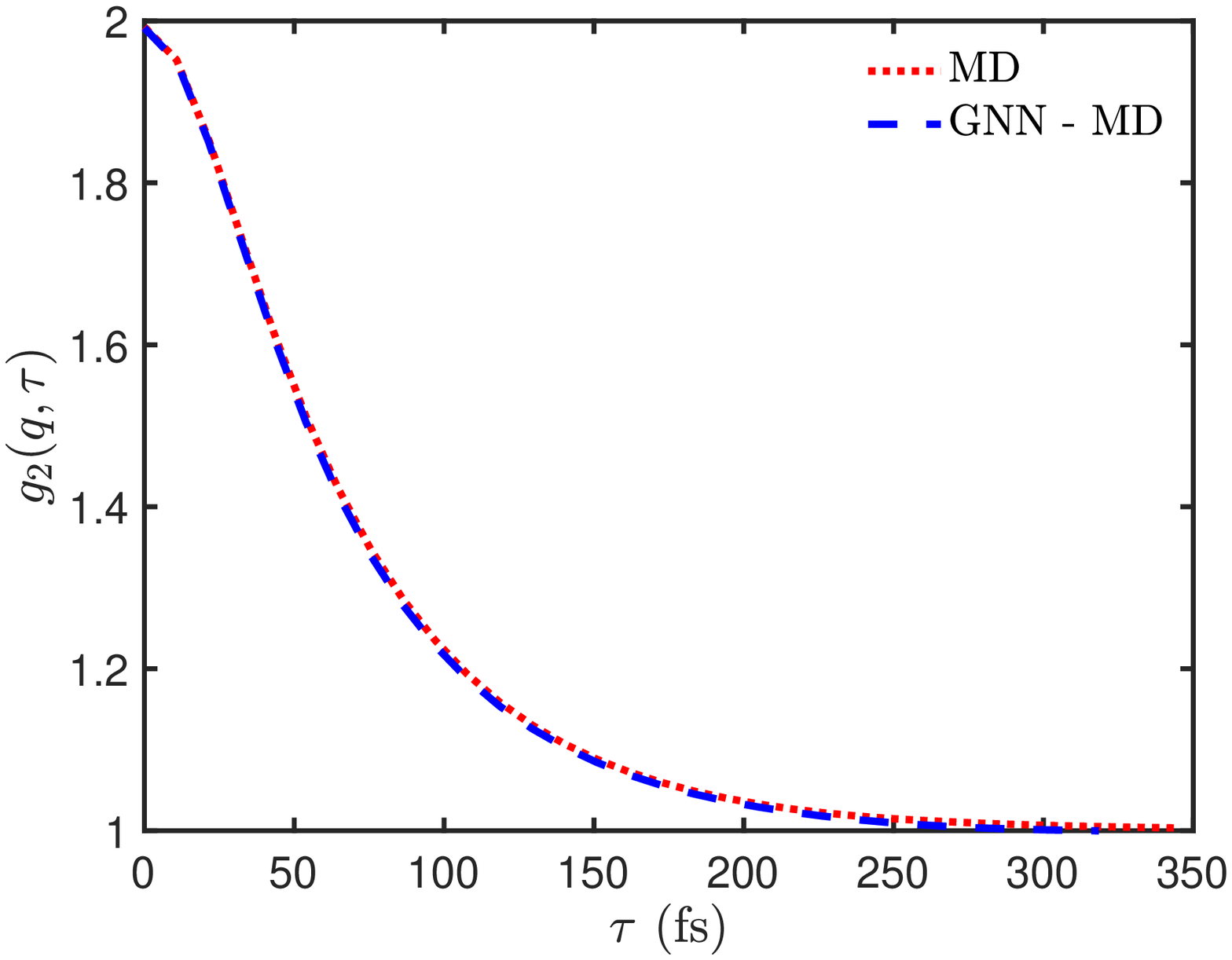}
    \label{fig:app-g2_4000}}
    \subfigure[]{\includegraphics[width=0.48\textwidth]{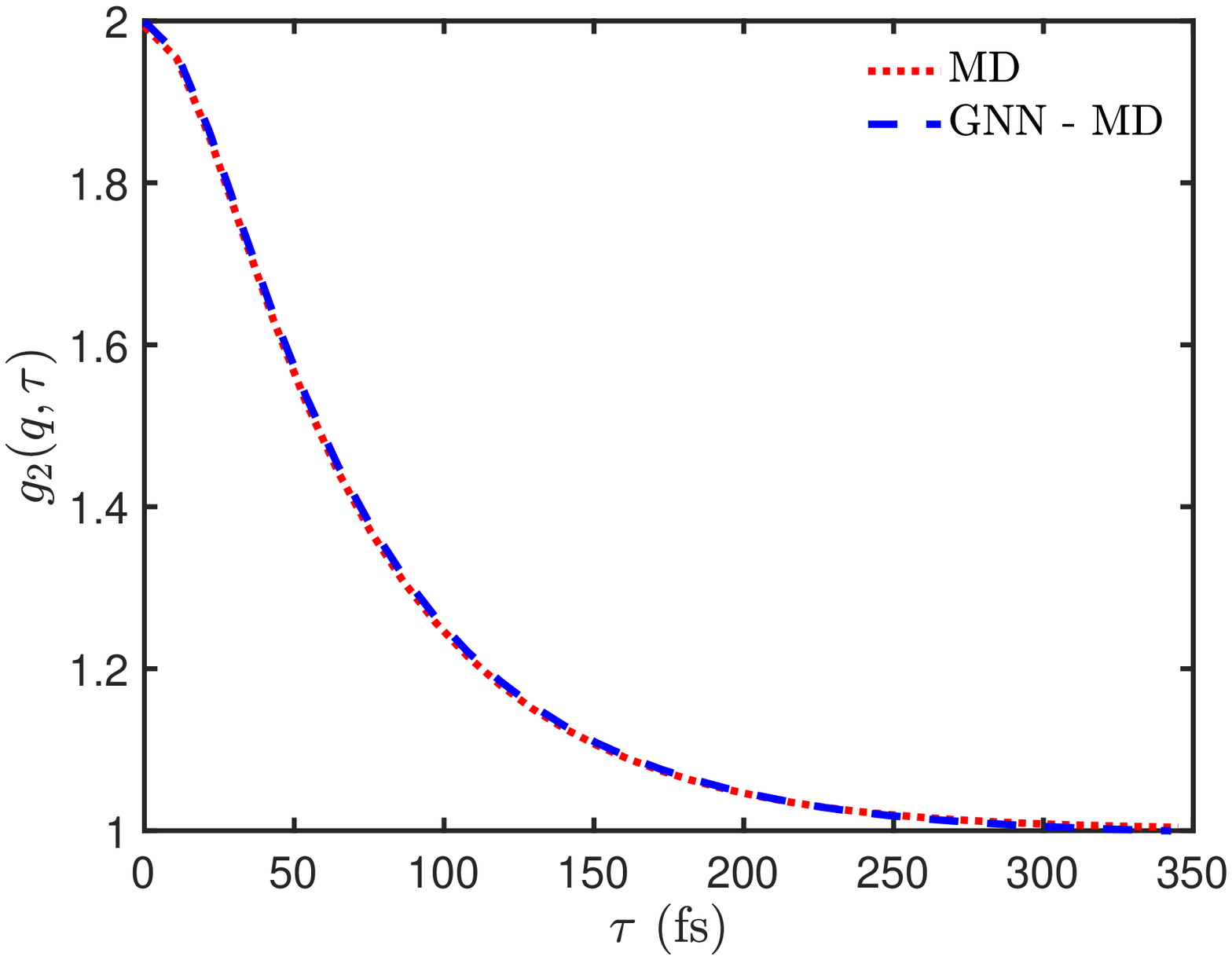}
    \label{fig:app-g2_95}}
    \subfigure[]{\includegraphics[width=0.48\textwidth]{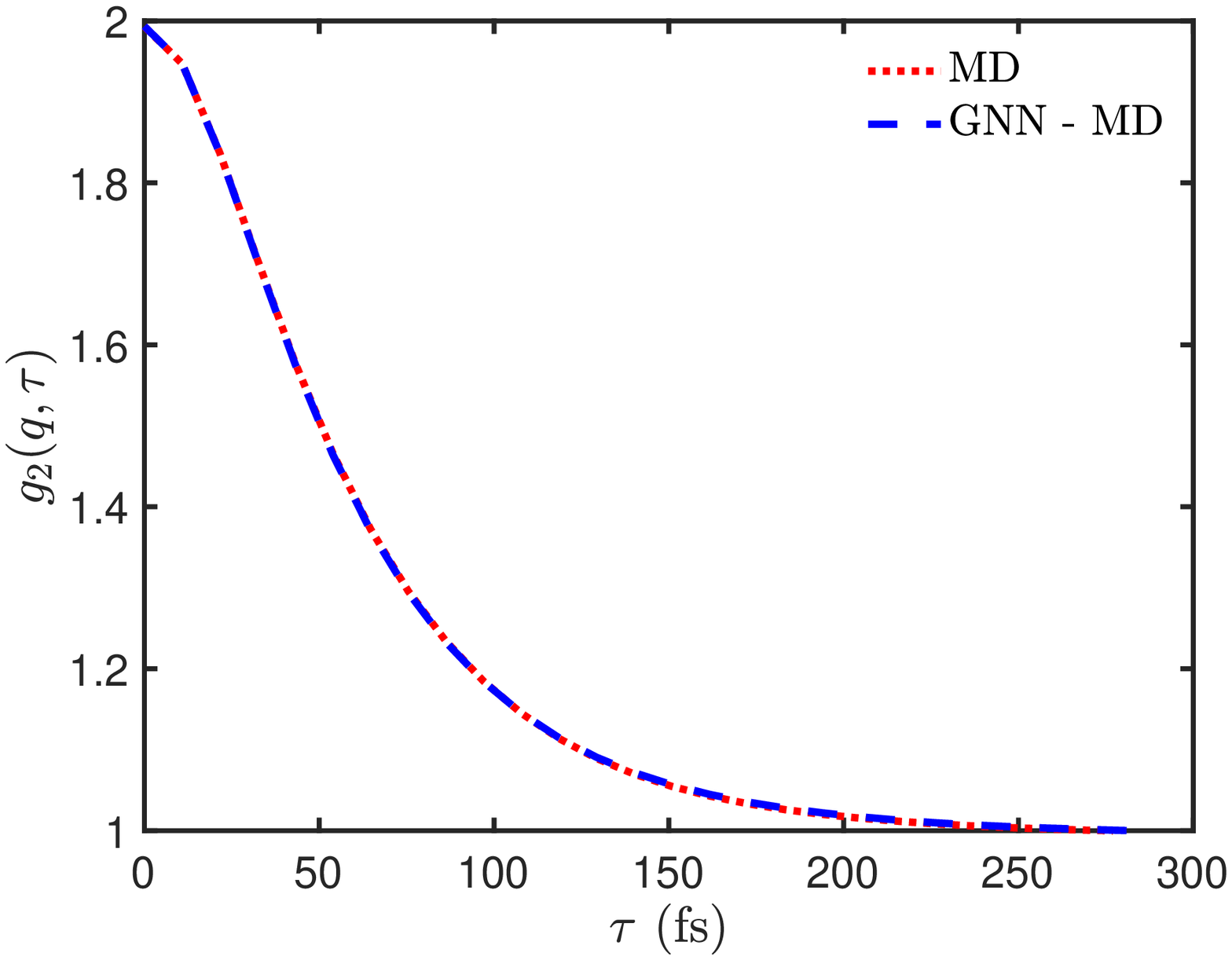}
    \label{fig:app-g2_110}}
    \caption{The $g_2(q,\tau)$ at $q = 1.844 \pm 0.029$ \AA$^{-1}$ over the first $\sim 3.56$ ps for the (a) 256 atoms (25 tracked atoms) at 100 K and the  4000 atoms (45 tracked atoms) at (b) 100 K, (c) 95 K and (d) 110 K, MD and GNN-MD simulation.  \label{fig:app-g2_all}}
\end{figure}
We observe that the decay in $g_2(q,\tau)$ between the MD and the GNN-MD simulation is also in agreement at all three temperatures.
\begin{figure}[H]
    \centering
    \subfigure[]{\includegraphics[width=0.48\textwidth]{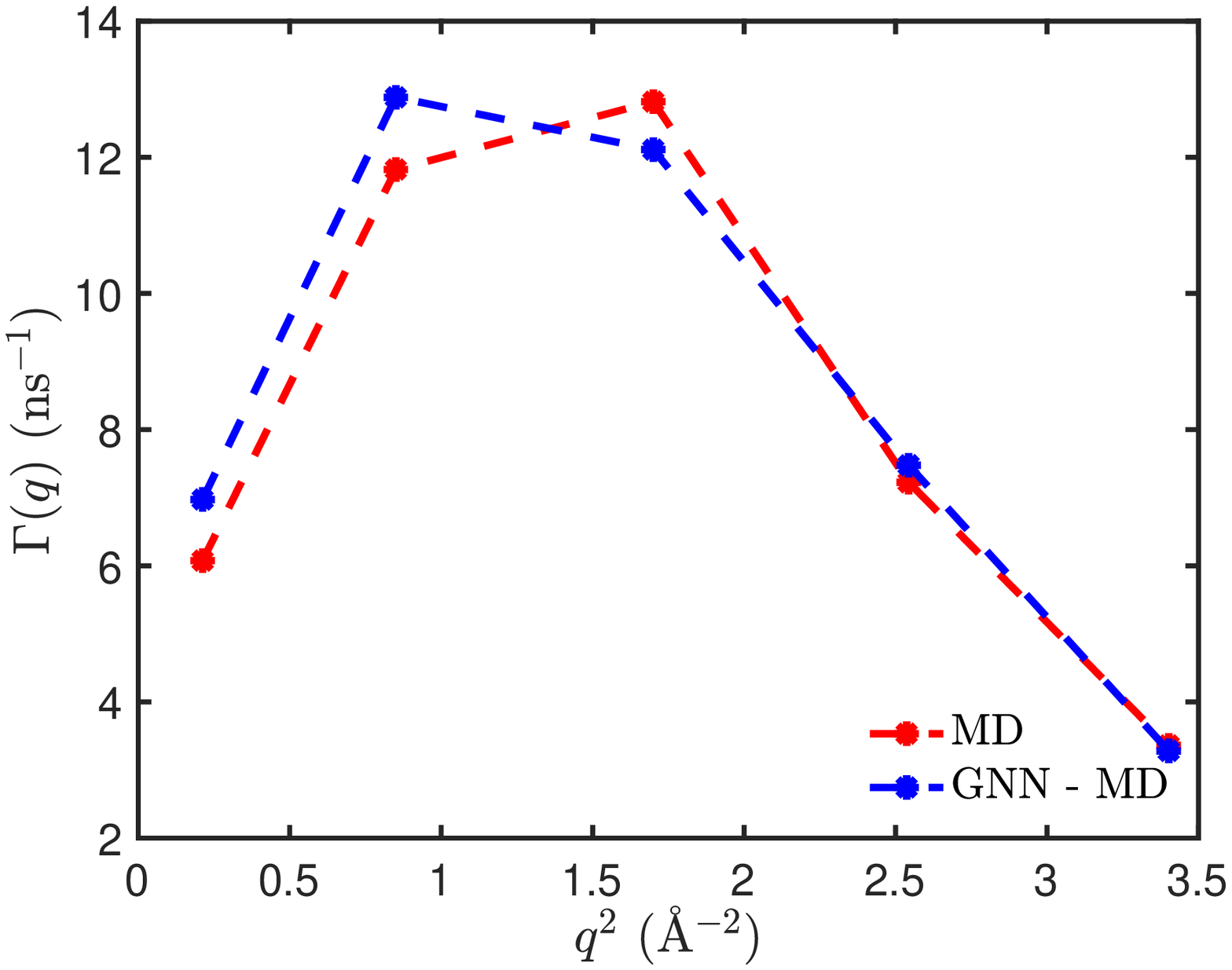}
    \label{fig:app-gamma_256}}
    \subfigure[]{\includegraphics[width=0.48\textwidth]{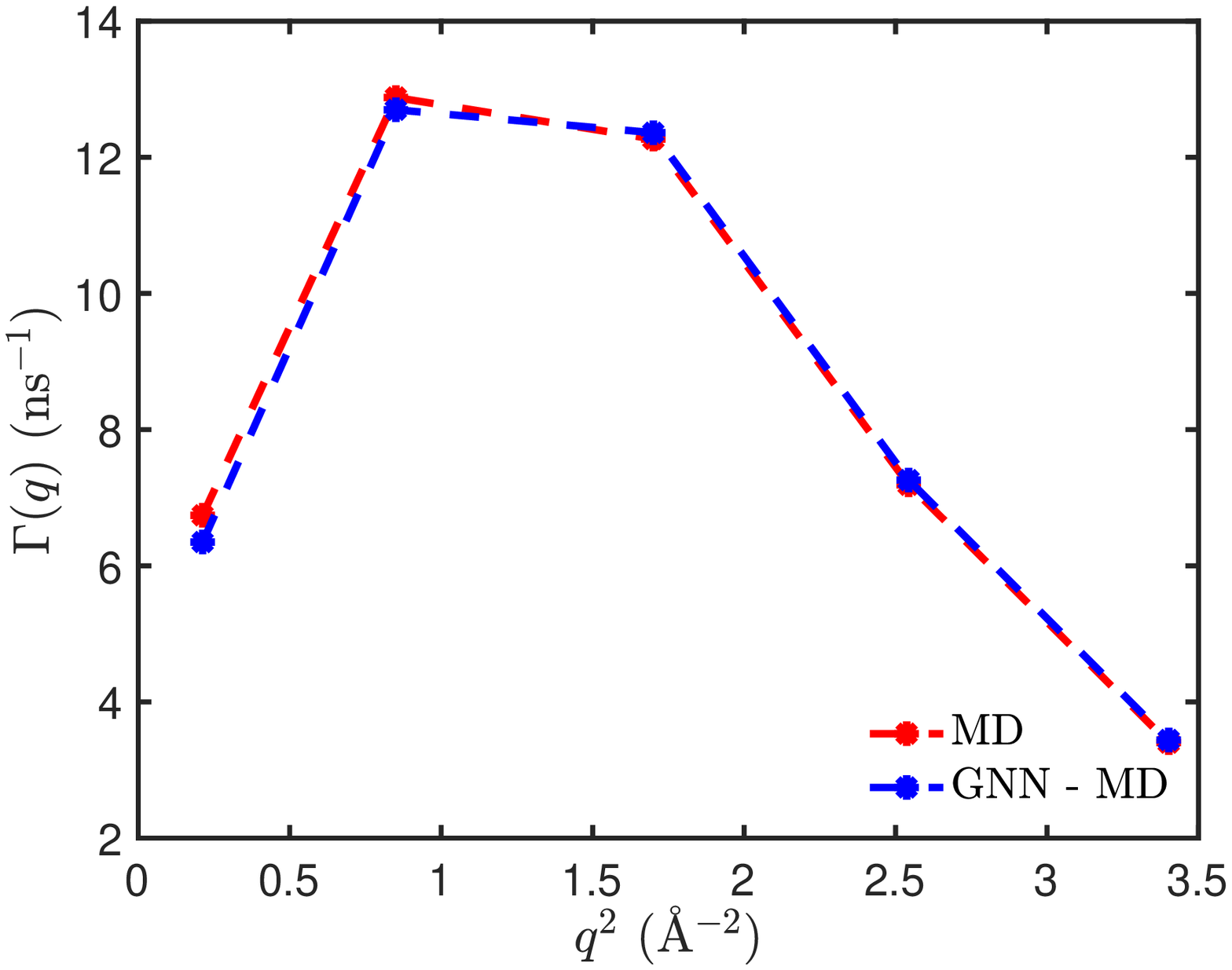}
    \label{fig:app-gamma_4000}}
    \subfigure[]{\includegraphics[width=0.48\textwidth]{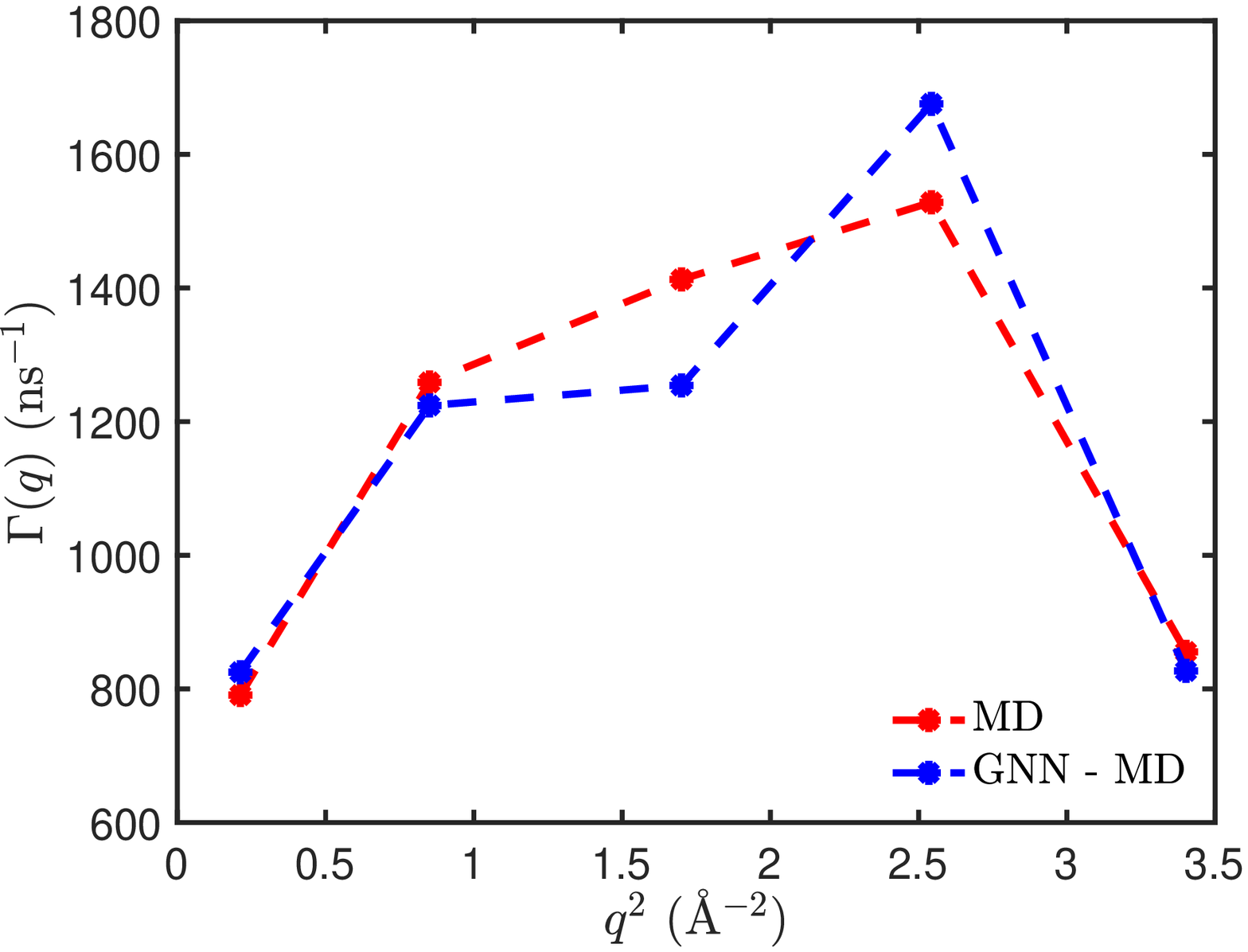}
    \label{fig:app-gamma_95}}
    \subfigure[]{\includegraphics[width=0.48\textwidth]{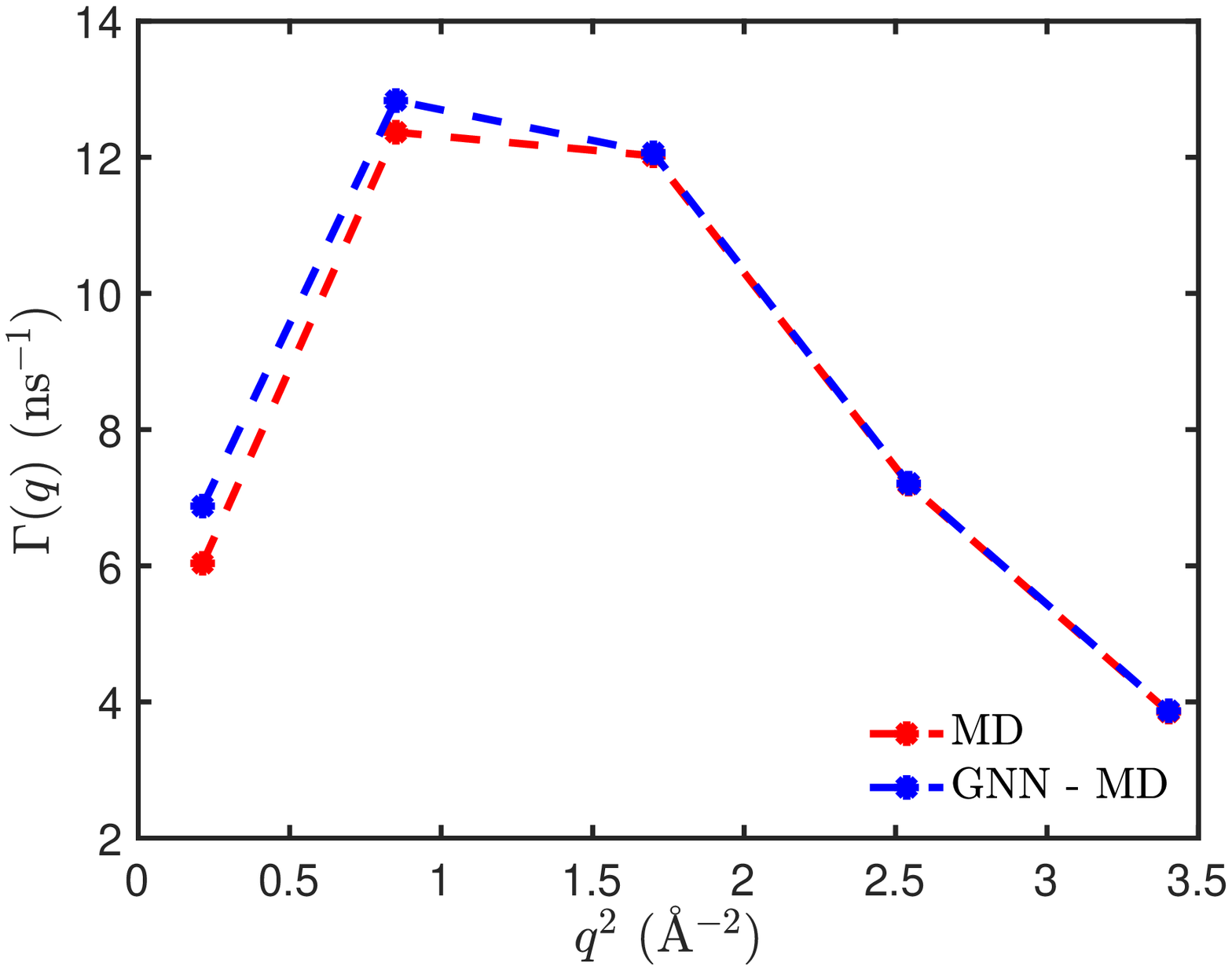}
    \label{fig:app-gamma_110}}
    \caption{The $\Gamma(q)$ as a function of $q^2$ for the (a) 256 atoms at 100 K and the  4000 atoms at (b) 100 K, (c) 95 K, and (d) 110 K, MD and GNN-MD simulation.  \label{fig:app-gamma_all}}
\end{figure}
We examine the $\Gamma(q)$ as a function of $q^2$ for different temperatures and see a very close agreement between the MD and GNN-MD simulations.
\begin{figure}[H]
    \centering
    \subfigure[]{\includegraphics[width=0.48\textwidth]{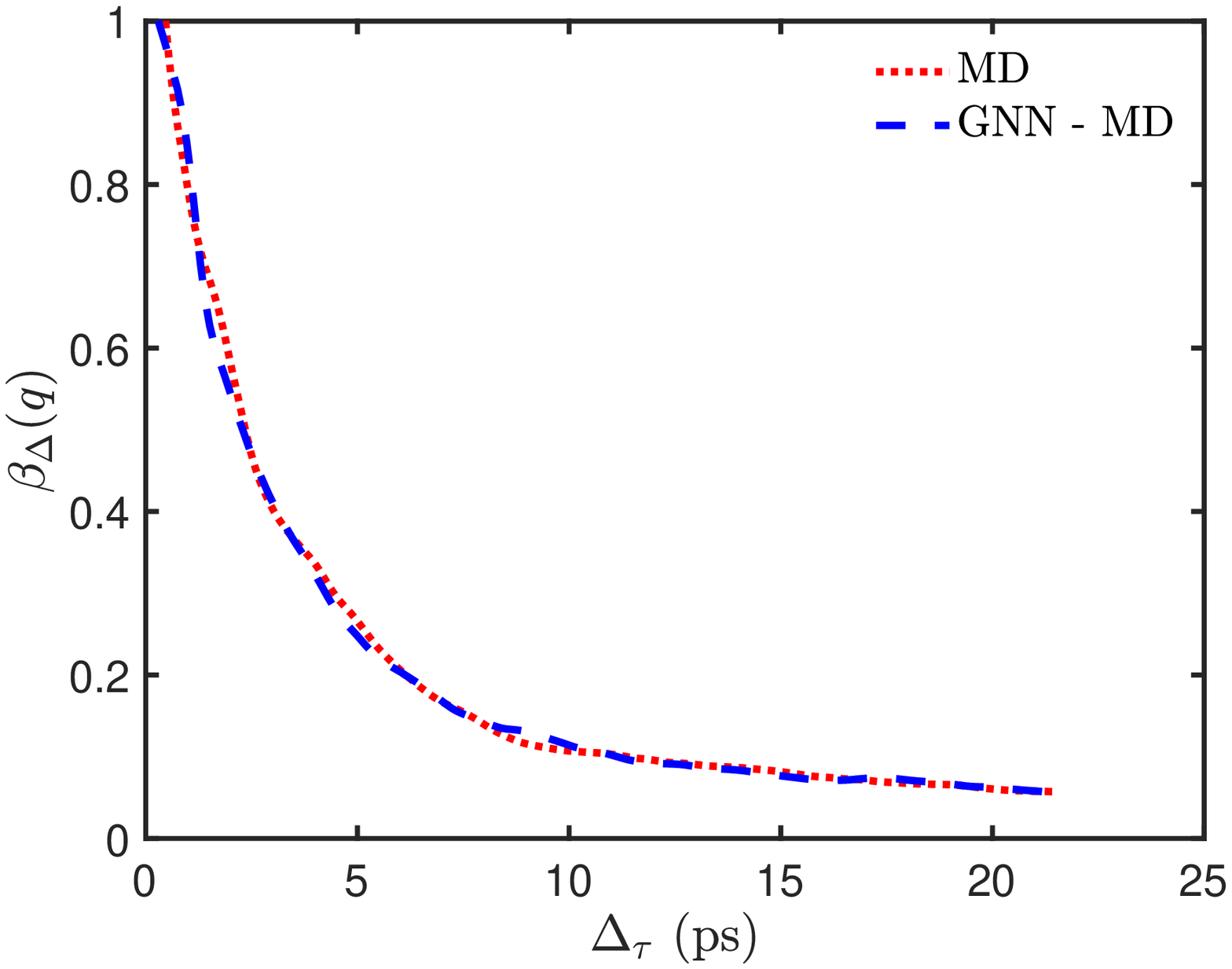}
    \label{fig:app-beta_256}}
    \subfigure[]{\includegraphics[width=0.48\textwidth]{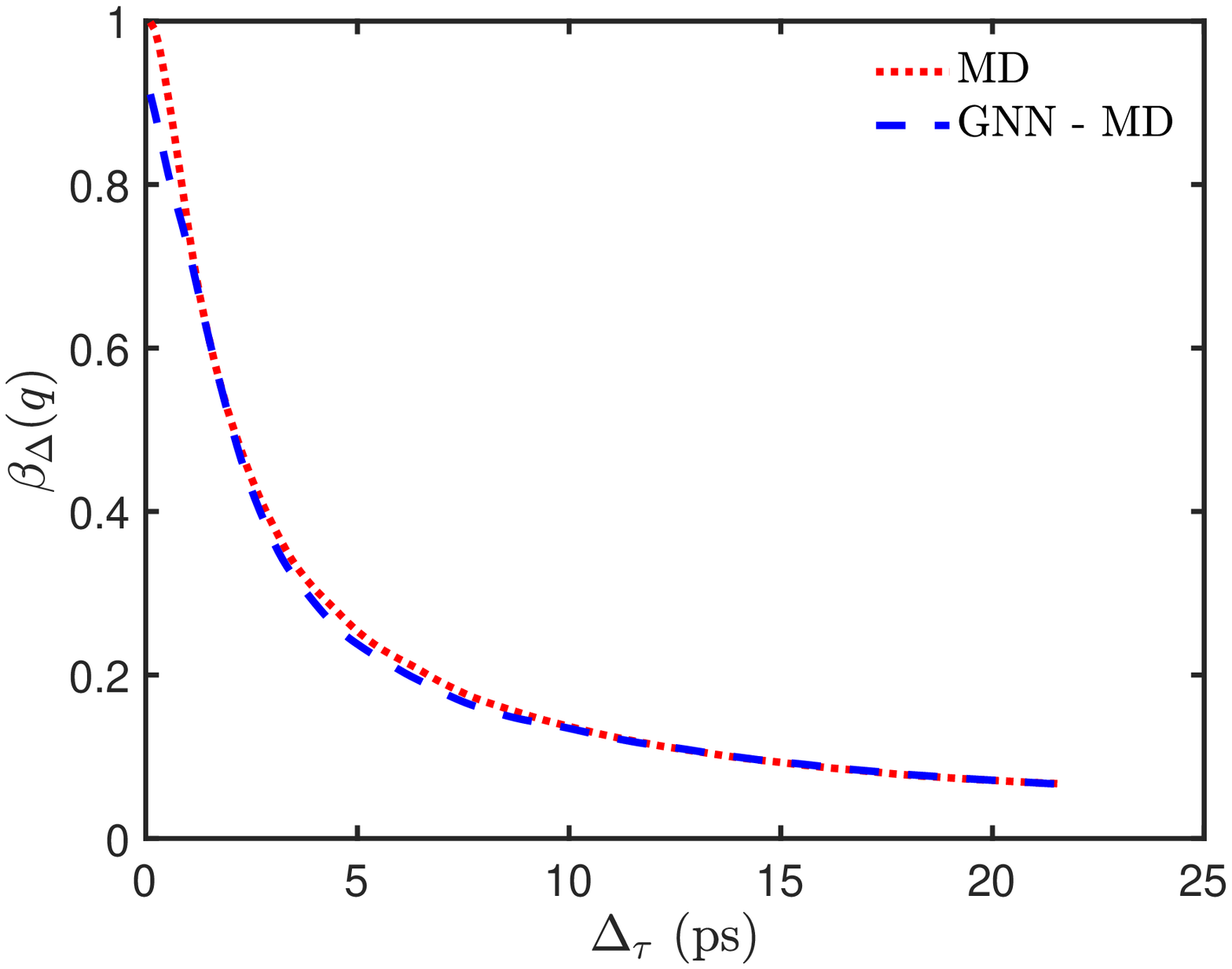}
    \label{fig:app-beta_4000}}
    \subfigure[]{\includegraphics[width=0.48\textwidth]{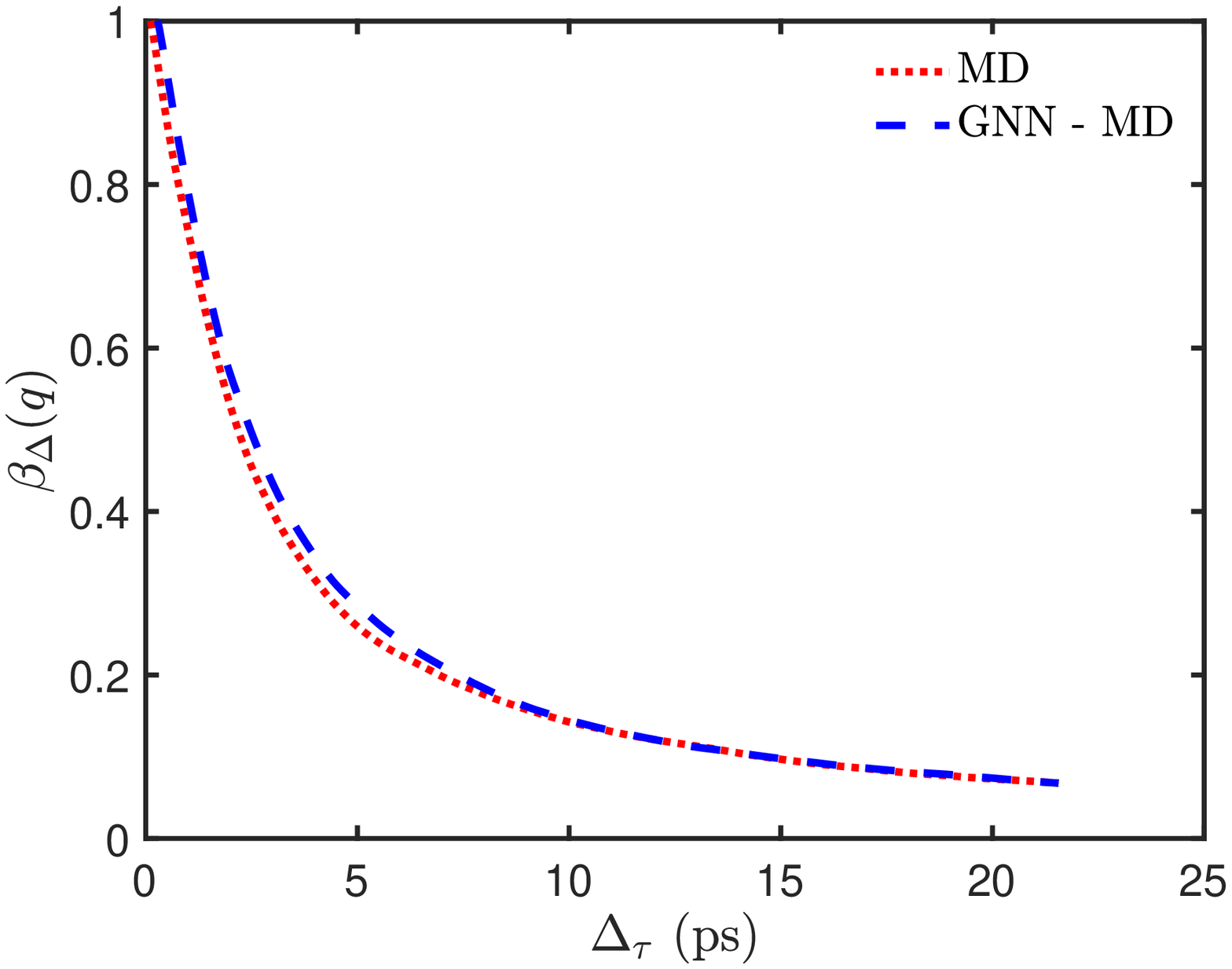}
    \label{fig:app-beta_95}}
    \subfigure[]{\includegraphics[width=0.48\textwidth]{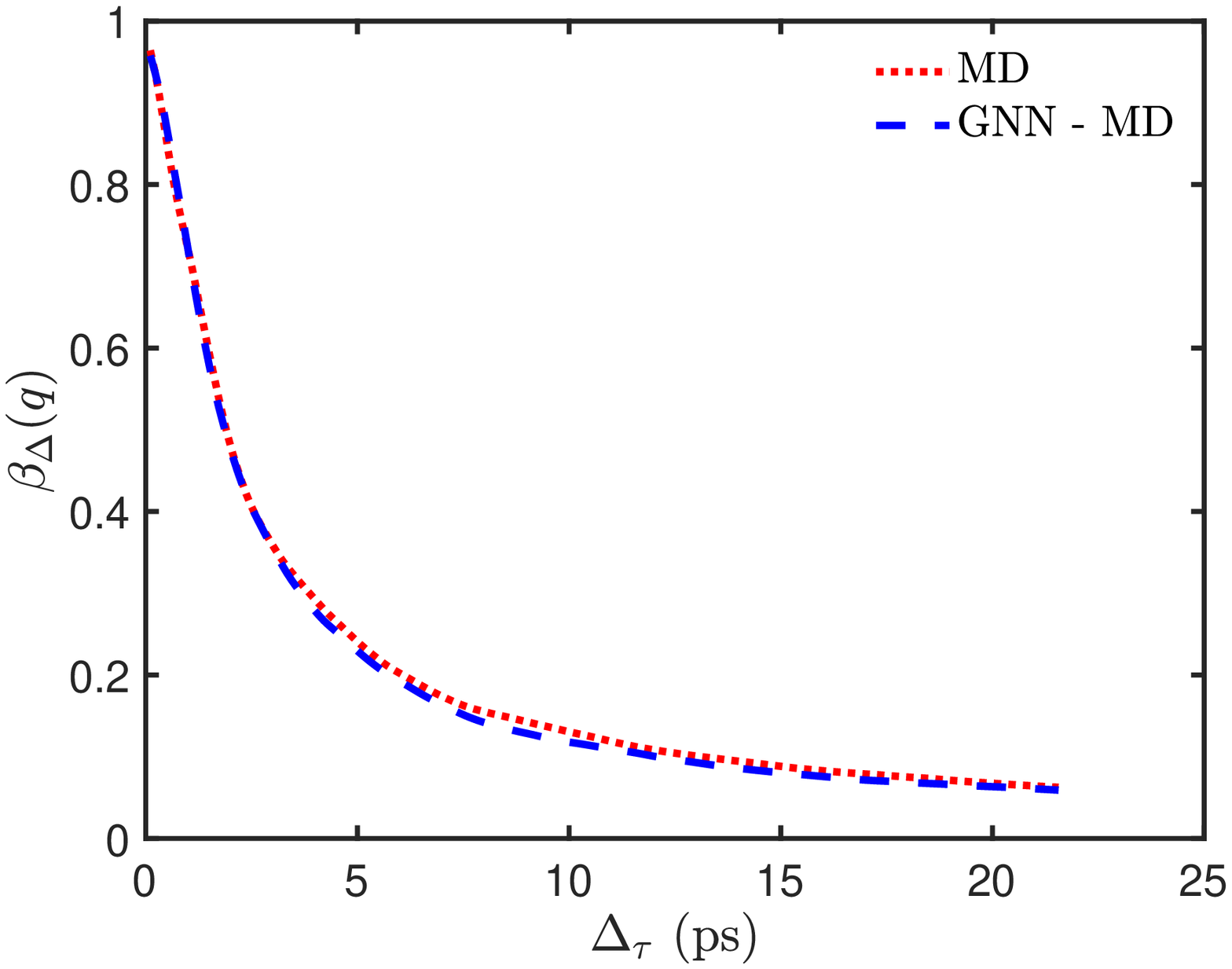}
    \label{fig:app-beta_110}}
    \caption{The $\beta_{\Delta}(q)$ as a function of $\Delta_{\tau}$ for the (a) 256 atoms at 100 K and the  4000 atoms at (b) 100 K, (c) 95 K and (d) 110 K, MD and GNN-MD simulation for $\lvert \bm{q} \rvert = 1.844 \pm 0.029$ \AA$^{-1}$.  \label{fig:app-beta_all}}
\end{figure}
In addition to the previous analysis, we also examine the XSVS analysis by plotting the optical contrast, $\beta_{\Delta}(q)$, as a function of the exposure time $\Delta_{\tau}$. The decay in the optical contrast is also in agreement irrespective of the temperature of the simulation which shows that the GNN force field approximates the Lennard-Jones potential sufficiently to capture the dynamics on liquid phase.
\end{document}